\documentclass[conference]{IEEEtran}

\usepackage[T1]{fontenc}
\usepackage{microtype}

\usepackage{color,soul} 
\usepackage{caption}
\usepackage{subcaption}
\usepackage{booktabs}
\usepackage{multirow}
\usepackage{algorithm}

\usepackage{siunitx}

\usepackage{makecell}

\usepackage{blindtext}

\usepackage{comment}

\usepackage{graphicx} 

\usepackage{amssymb}
\usepackage{listings} 
\usepackage{verbatim}
\usepackage{booktabs}
\usepackage{float}

\usepackage{tabularx} 
\usepackage{xcolor}

\usepackage{amsmath}

\usepackage{amsmath,amssymb,amsfonts}
\usepackage{algorithmic}
\usepackage{graphicx}
\usepackage{textcomp}
\usepackage{placeins}

\IEEEoverridecommandlockouts

\usepackage{todonotes} 
\let\oldtodo\todo
\renewcommand{\todo}[1]{\oldtodo[inline]{#1}}

\usepackage{cite}
\usepackage{amsmath,amssymb,amsfonts}
\usepackage{algorithmic}
\usepackage{graphicx}
\usepackage{textcomp}
\usepackage{xcolor}

\begin{document}

\title{
ArrythML: An Autoencoder-Based TinyML Approach for On-Device Arrhythmia Detection on Resource-Constrained Embedded Systems}

\author{
    Nagarajan S, 
    and Kurian Polachan,~\IEEEmembership{Senior Member,~IEEE}%
    \\
    \textit{International Institute of Information Technology, Bangalore, India} \\
    \{nagarajan.s, kurian.polachan\}@iiitb.ac.in

\thanks{
All authors are with the International Institute of Information Technology, Bangalore, India. Corresponding author: Kurian Polachan (email: kurian.polachan@iiitb.ac.in).

An earlier version of this paper was presented at the IEEE 7th International Conference on Smart Applications, Communications and Networking (SmartNets), Istanbul, Turkiye, 2025~\cite{nagarajan2025autoencoder}}

}

\maketitle

\begin{abstract}

Our work presents a method for ECG segmentation and arrhythmia detection using Tiny Machine Learning (TinyML) models for real-time, on-device inference on resource-constrained embedded systems. We develop INT8 quantized  autoencoder-based TinyML models with minimal layers and parameters for embedded deployment. 
These models are evaluated using a custom dataset derived from the MIT-BIH Arrhythmia Database and validated in both PC-based simulations and on-device environments.
For the evaluations, over 95,000 ECG segments are processed on an ESP32-S3 microcontroller running the TensorFlow Lite Micro runtime. Post-evaluation, detailed analysis, including annotation-wise and record-wise failure analysis is conducted to characterize model behavior across diverse ECG morphologies and rhythm patterns and to explain missed detections. In several cases, apparent misclassifications may correspond to early or subtle anomaly patterns labeled as normal in the reference annotations, highlighting the model’s sensitivity. A refined evaluation by filtering out ambiguous cases in the  dataset shows that the best-performing DNN-based autoencoder achieves a recall of 84\%, an F1-score of 79\%, a model size of approximately 180 KB, and an inference latency of 9~ms on-device. These results demonstrate the feasibility of low-power, privacy-preserving embedded wearable systems capable of performing accurate arrhythmia detection entirely on-device.

\end{abstract}


\section{Introduction}
\label{sec_introduction}
An electrocardiogram (ECG) measures the electrical activity of the heart and is an important tool for monitoring and diagnosing heart conditions, such as arrhythmia~\cite{Moody2001}. Continuous, real-time ECG monitoring helps detect  early signs of arrhythmia~\cite{kakria2015realtime}, enabling timely medical intervention to reduce the risk of life-threatening conditions. However, current ECG analysis techniques are often manual and require hospital visits, trained professionals and lengthy waiting periods for results, all of which can delay timely diagnosis~\cite{Rangayyan2002}.

In recent years, ECG monitoring devices have gained popularity by enabling heart health monitoring outside of the hospital setting \cite{Rangayyan2002}. These wearable devices record ECG signals for a specified duration and then transfer the recordings to a nearby hub (e.g., a mobile phone or PC) or to a cloud server for post-processing, where machine learning (ML) algorithms are used to detect abnormalities. However, these devices suffer from several shortcomings. First, they require frequent wireless transmission of ECG recordings, and wireless communication via BLE or WiFi consumes a significant amount of energy~\cite{agarwal2024vlc}. This energy consumption reduces battery life, often forcing users to recharge or replace batteries frequently. Moreover, wireless data transfer raises the risk of interception; a nearby attacker could potentially capture the transmitted signals and decode sensitive personal health information~\cite{majumdar2025hbclock, jagan2025murmur}.  Although encrypting wireless transfers is a possibility, securing the encryption key and encryption process against potential attacks requires implementing security techniques, which adds to the energy consumption and latency of these wearables~\cite{vaidya2020gpio, chakrabarty2024iopuf, naz2024led}. Additionally, offloading ECG processing to a hub or cloud adds delays in the diagnosis of arrhythmias.

An alternative approach is to develop wearable devices that record and process ECG signals locally. Rather than transmitting ECG data for external processing, these devices themselves could run ML algorithms to detect abnormalities such as arrhythmias. This approach reduces energy consumption and improves data security by avoiding frequent wireless transmissions. Further, they speed up the diagnostic process, enabling quicker medical interventions. However, implementing ML models locally presents a challenge: many wearable devices utilize microcontroller units (MCUs). MCUs are significantly more cost-effective and consume less energy. However, MCUs are significantly resource-constrained, their processing power and memory capacity (both flash and RAM) are considerably lower, making it difficult to run traditional ML models like deep neural networks (DNNs) that require high-end computing hardware.

Our work aims to address these challenges by exploring Tiny Machine Learning (TinyML) algorithms for detecting arrhythmias that fit within the limited flash and RAM space of MCUs. Among the many different ML architectures in the literature, we selected autoencoder-based ML architecture in this work for its ability to detect anomalies and provide interpretability of its predictions. An autoencoder follows an encoder–decoder architecture in which ECG segments fed are first compressed into a lower-dimensional latent representation and subsequently reconstructed to approximate the original input. The model is trained exclusively on normal ECG segments to minimize reconstruction error. During inference, normal ECG segments produce low reconstruction error, whereas abnormal segments, such as those corresponding to arrhythmias, result in significantly higher reconstruction error. Thresholding the reconstruction error helps classify the input ECG segment as normal or abnormal. Beyond such quantitative classification, the autoencoder architecture also helps us, as designers of the model, to inspect the correctness of the ML predictions. Since the reconstructed ECG signal from the autoencoder corresponds to the model’s learned expectation of normal cardiac activity, a visual comparison of this with the input ECG segment helps qualitatively identify portions of the waveform that deviate from normal behavior. Such qualitative probing of results to understand the rationale behind a prediction is not inherently available in many conventional black-box classification models.

In this work, we investigate two lightweight autoencoder variants for ECG anomaly detection: a DNN-based autoencoder and a CNN-based autoencoder~\cite{berahmand2024autoencoders}. Both architectures are designed to be compact, employing a minimal number of layers and neurons sufficient to achieve acceptable detection performance. Furthermore, their model parameters are quantized from floating-point representation to 8-bit integers (INT8) to reduce memory footprint and inference latency for improved hardware performance. To mitigate potential degradation in prediction accuracies due to quantization, quantization-aware training (QAT) is employed, where the model parameters are trained with simulated INT8 precision. We evaluate the models both in a PC-based simulation environment and on ESP32-S3-based hardware using a custom engineered dataset derived from the MIT-BIH Arrhythmia Database~\cite{MITBIHDatabase}.

Over 95,000 ECG segments are processed on-device using an ESP32-S3 microcontroller running the TensorFlow Lite Micro runtime during evaluations. Post-evaluation, detailed record-wise failure analysis is performed to characterize model behavior across diverse ECG morphologies and rhythm patterns, and to explain missed detections. In several cases, this analysis reveals that apparent misclassifications correspond to early or subtle arrhythmic patterns labeled as normal in the reference annotations, highlighting the model’s ability to flag segments at the onset of arrhythmia. In some instances, misclassifications are also attributed to the presence of noise in the ECG signal.

A refined evaluation of the models is therefore carried out by excluding problematic records and segments identified during the failure analysis, resulting in a more representative performance assessment. These evaluation results indicate that the DNN-based autoencoder achieves strong recall and balanced overall performance, while maintaining a compact model size and low inference latency on hardware. These results demonstrate the feasibility of developing low-power, privacy-preserving wearable systems capable of performing on-device arrhythmia detection without reliance on external hubs or cloud infrastructure.

\subsection{Contributions}

This paper makes the following contributions:

\begin{itemize}
\item It develops an R–R interval–based ECG segmentation method and quantized, lightweight CNN/DNN-based autoencoder models for on-device arrhythmia detection on low-power, resource-constrained wearable platforms.

\item It characterizes the performance of quantized models on microcontroller hardware using a large evaluation dataset of approximately 95,000 ECG segments derived from the MIT-BIH Arrhythmia Database.

\item It provides detailed performance analysis, including record-wise failure analysis on the MIT-BIH Arrhythmia Database to identify sources of false positives and false negatives, and outlines insights and future directions applicable to quantized models for healthcare deployment.
\end{itemize}

\subsection{Outline}

The remainder of this paper is organized as follows. Section~\ref{sec_related_work} reviews related work in the literature. Section~\ref{sec_system_design} presents our proposed system for real-time ECG segmentation and arrhythmia detection using autoencoder-based ML models. The section also discusses the training of these models, their quantization, and deployment on hardware. Section~\ref{sec_results} presents the experimental setup along with the results; characterizing the performance of autoencoder-based ML models in simulation and on hardware. Section~\ref{sec_discussion} discusses the results. Finally, Section~\ref{sec_conclusion} concludes the paper by summarizing the key findings and providing suggestions for future work.

\section{Related Work}
\label{sec_related_work}

Automated analysis of ECG signals has been extensively studied for the detection of cardiac abnormalities, particularly arrhythmias. Existing literature in this domain can be broadly grouped into three categories based on the analysis techniques employed: classical signal-processing-based approaches, high-capacity deep-learning-based techniques, and techniques involving model compression for deployment at the edge.

\subsection{Classical Signal-Processing-Based Methods}

Early research on automated ECG analysis primarily relied on rule-based signal-processing techniques. One of the most widely adopted approaches in this domain is the Pan--Tompkins algorithm, which performs QRS detection through a sequence of filtering, differentiation, squaring, and moving-window integration stages, followed by adaptive thresholding to identify R-peaks~\cite{pan1985real}. Subsequent studies focused on improving the robustness and accuracy of QRS detection algorithms by leveraging annotated ECG datasets, particularly the MIT-BIH Arrhythmia Database~\cite{hamilton1986quantitative}. The availability of annotated ECG datasets enabled the systematic evaluation and refinement of detection rules under diverse physiological and noise conditions.

Research efforts also expanded toward extracting features beyond QRS to enable more comprehensive ECG analysis. This included the identification of waveform boundaries (P, QRS, and T waves) and the extraction of clinically relevant temporal parameters such as QRS duration and QT intervals~\cite{laguna1994new}. These works enabled the development of more sophisticated rule-based ECG analysis methods that leverage a richer set of morphological and temporal characteristics beyond those used in earlier QRS-centric approaches for arrhythmia detection. Such rule-based approaches formed the backbone of early clinical monitoring systems, including bedside cardiac monitors and Holter analysis platforms. These systems were commonly implemented using dedicated ECG recording hardware along with often bulky workstations running rule-based software for automated and continuous cardiac monitoring.

Despite their widespread adoption, rule-based signal-processing approaches depend heavily on handcrafted features and manually designed decision rules. Consequently, their performance degrades in the presence of noise, varies across patients, and struggles to detect complex arrhythmia patterns~\cite{clifford2006advanced}. These limitations have motivated researchers to explore machine-learning-based approaches that can learn ECG morphology directly from input ECG samples, leading to improved performance.

\subsection{Compute-Intensive Deep Learning Methods}

Recent advances in machine learning have enabled new approaches for automated ECG analysis. With the emergence of deep learning techniques, numerous studies have employed deep neural network (DNN) and convolutional neural network (CNN) architectures for arrhythmia detection and classification~\cite{acharya2017deep,rajpurkar2017cardiologist,jun2018ecg}. These models learn hierarchical feature representations directly from raw ECG waveforms, thereby eliminating the need for handcrafted features used in traditional signal-processing approaches.

Deep convolutional and hybrid CNN--RNN architectures have demonstrated strong performance on benchmark datasets such as the MIT-BIH Arrhythmia Database. For instance, Rajpurkar \textit{et al.} and Hannun \textit{et al.} have reported very deep convolutional architectures capable of identifying multiple arrhythmia types with performance comparable to cardiologists~\cite{rajpurkar2017cardiologist,hannun2019cardiologist}. In addition, patient-specific deep learning approaches have been explored, where models are trained or adapted for individual patients~\cite{kiranyaz2016real}.


Despite their strong performance, deep learning models for ECG analysis are typically compute-intensive, often involving millions of parameters and high memory requirements. Consequently, these models are generally evaluated on workstation- or cloud-based platforms with substantial computational resources~\cite{rajpurkar2017cardiologist,acharya2017deep,kiranyaz2016real}. Deployment of these models on wearable devices is challenging due to the resource-constrained nature of embedded systems, where energy efficiency, memory capacity, and processing capability are limited.

\paragraph*{Note}
Issa \textit{et al.}~\cite{Issa2023} presented a deep learning-based heartbeat classification approach using single lead-II ECG signals. Although single lead-II ECG signals are often used in a wearable setting, the study focuses on running ML algorithms in  conventional computing environments, the study does not address constraints associated with embedded deployment of ML models. Further, the work provides limited discussion on dataset coverage and ectopic beat handling.

\subsection{Model Compression and Edge-Oriented Approaches}

In response to the deployment constraints of deep learning models on embedded platforms, recent research has explored techniques such as quantization and pruning to reduce model size. These approaches help lower computational complexity, memory requirements, and energy consumption during inference, making models more suitable for resource-constrained hardware.

However, many of these works still target embedded edge-AI platforms built around relatively powerful microprocessors or specialized accelerators such as ASICs and FPGAs. Such platforms offer significantly higher computational capability and memory compared to microcontroller-based systems, which are characterized by limited on-chip memory, reduced numerical precision, and stringent energy budgets~\cite{banbury2021benchmarking}, as commonly found in wearable devices.

Only limited work has focused on designing ECG analysis methods specifically tailored for low-power microcontroller platforms used in wearable devices. Preliminary investigations in this direction were presented in our earlier work~\cite{nagarajan2025autoencoder}, where a lightweight autoencoder model was proposed to analyze segmented ECG signals and detect deviations from normal cardiac activity. However, that study primarily focused on proof-of-concept validation and did not comprehensively optimize model behavior for deployment, nor did it evaluate performance on actual hardware.

The present work significantly extends our prior study~\cite{nagarajan2025autoencoder} in several aspects. First, a deployment-aware autoencoder framework for low-power, on-device arrhythmia detection is developed, incorporating a convolutional (CNN-based) autoencoder alongside the previously explored fully connected (DNN-based) model. Second, the models are quantized using quantization-aware training, along with dataset refinements and model fine-tuning to improve sensitivity (recall) and overall classification accuracy. Finally, the proposed approach is evaluated on both PC-based platforms and an ESP32-S3 microcontroller using ECG segments derived from the MIT-BIH Arrhythmia Database, where the majority of records are utilized for evaluation, excluding only those with significant signal quality issues. The evaluation is further extended to include record-wise failure analysis to provide deeper insights into model behavior under diverse ECG records.

\paragraph*{Note}
Recently, wearable-based atrial fibrillation (AFib) detection methods have also been reported in the literature. These approaches commonly rely on R--R interval irregularity analysis using rule-based or statistical techniques rather than machine learning models~\cite{Tateno2001AF, Lian2011AF}. Such methods typically analyze inter-beat interval variability derived from ECG or photoplethysmography (PPG) signals and are mainly targeted toward detecting specific rhythm abnormalities, including bradycardia, tachycardia, irregular rhythms, and heart rate variability. In contrast, ML-based approaches can generalize to a broader range of arrhythmia detection tasks by learning complex morphological and temporal ECG patterns.

\section{System Design}
\label{sec_system_design}

The first part of this section presents the proposed conceptual system for on-device arrhythmia detection using R–R segmentation of the ECG signal and an autoencoder-based ML model. The second and the third part discusses the rationale behind adopting R--R segmentation and choosing an autoencoder architecture, respectively. The final part describes two variants of autoencoder models explored in this work, along with their development processes and deployment workflows on an embedded platform.

\subsection{On-Device Autoencoder-Based Arrhythmia Detection}
\label{sec_sys_design_arrythmia_detection}
\begin{figure*}[!htbp]
    \centering
    \includegraphics[width=1\linewidth]{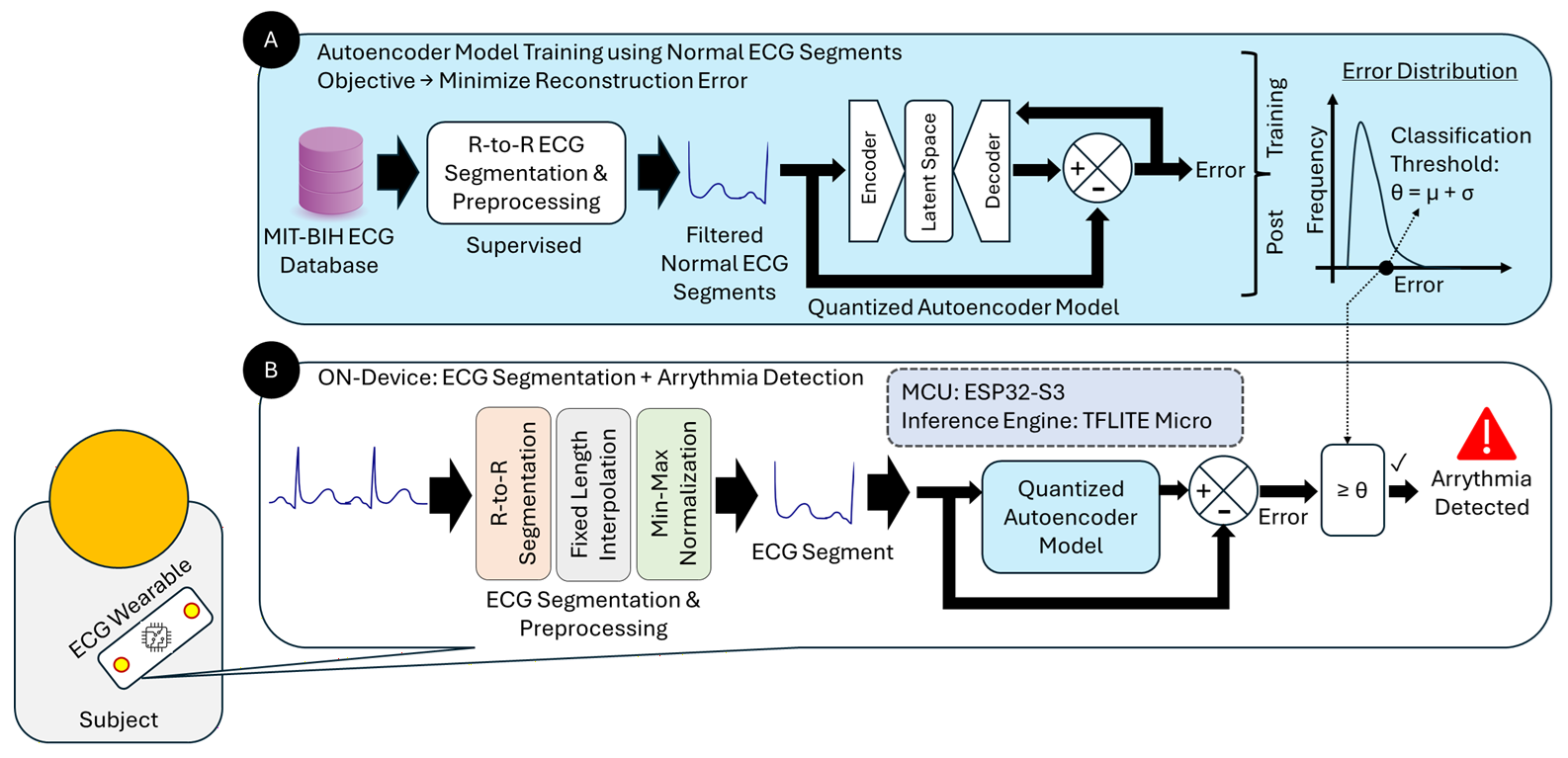}
    \caption{(a) Autoencoder training process utilizing multiple patient records from the MIT-BIH Arrhythmia Database. (b) Conceptual diagram showing on-device real-time arrhythmia detection using an autoencoder-based machine learning model.
    }
    \label{fig_system_design}
\end{figure*}

Fig.~\ref{fig_system_design}~(a) shows our proposed conceptual system for on-device arrhythmia detection. The system consists of a subject wearing an ECG patch that records ECG signals and runs an autoencoder-based machine learning (ML) model to detect arrhythmias in real time. The wearable device continuously records ECG signals, which are then segmented (R-to-R) in real time and fed into a pre-trained autoencoder model for detecting arrhythmias. 

An autoencoder is a type of neural network that follows an encoder-decoder architecture~\cite{goodfellow2016deep}. When an ECG segment is fed into the model, the encoder compresses it into a lower-dimensional latent space. Then, the decoder reconstructs the original ECG segment from this compressed representation. 

During training, the autoencoder is fed normal ECG segments extracted from multiple patient records in the MIT-BIH ECG database. See Fig.~\ref{fig_system_design}~(b). The model weights and biases are optimized during training to minimize the reconstruction error, measured as the mean squared error between the reconstructed and original ECG segments. 

Post-training, when the autoencoder is fed normal ECG segments, it generates low reconstruction errors, as it has been trained on similar normal ECG segments. However, when fed abnormal ECG segments characterized by arrhythmias, the model generates high reconstruction errors. 

To determine the reconstruction error threshold for detecting arrhythmias, we analyze the reconstruction error distribution post-training. The autoencoder is fed normal ECG segments, and the reconstruction errors are recorded. From the distribution of these reconstruction errors, the mean ($\mu$) and standard deviation ($\sigma$) are calculated. We set the classification threshold as $\theta = \mu + \sigma$. During inference, if the reconstruction error of an ECG segment is below this threshold, it is classified as normal. Otherwise, it is labeled as abnormal, indicating a potential arrhythmia.

\subsubsection{Rationale for Adopting R-to-R Segmentation}

In an ECG signal, the PQRST complex represents the electrical activity of the heart over a cardiac cycle. The P wave corresponds to atrial depolarization, the QRS complex reflects ventricular depolarization, and the T wave represents ventricular repolarization~\cite{ecgwaves_ecg_basics}. Among these features, the R-to-R interval is widely used to estimate heart rate, typically expressed in beats per minute (bpm)~\cite{clifford2006advanced}. Consequently, most commercial wearable devices primarily focus on detecting R-peaks and measuring the interval between successive R-peaks for heart rate estimation.

Despite the significance of R-peak identifications for heart rate estimations, they are less commonly used for arrhythmia detection. Many prior studies on arrhythmia detection rely on identifying QRS complexes ~\cite{pan1985real}, and segmenting ECG around them for arrhythmia detection. These approaches rely on compute-intensive Digital Signal Processing (DSP) techniques to delineate the onset and offset of the complex accurately. 

However, such high-overhead processing is often incompatible with the wearable platforms often built using resource constrained low-cost commercial off-the-shelf (COTS) microcontroller units (MCUs). These MCUs often lack the computational resources and DSP capabilities required for robust QRS detection. As a solution, in this work, rather than segmenting ECG by detecting QRS complexes, this work carries out R-to-R segmentation by detecting R-peaks in the ECG. R-peak detection and subsequent R-to-R segmentation can be implemented using simple thresholding and lightweight filtering techniques, making them well-suited for implementation on MCUs.

\subsubsection{Rationale for Autoencoder Choice}


As discussed in Sections~\ref{sec_introduction}~and~\ref{sec_sys_design_arrythmia_detection}, autoencoder (AE)-based models classify ECG segments as normal or abnormal using reconstruction error. The model is trained to learn the characteristics of normal ECG waveforms, and abnormalities are identified as deviations between the input and reconstructed signals. This reconstruction-based approach is particularly effective for detecting subtle morphological variations in ECG signals. Unlike conventional classifiers that rely on predefined decision boundaries, AEs identify anomalies as deviations from learned normal patterns. Consequently, they can detect waveform variations that may closely resemble normal beats while still exhibiting minor structural abnormalities. The reconstruction error also provides an interpretable indicator of model behavior by highlighting waveform regions where significant deviations occur between the input and reconstructed output. Such behavior can assist in identifying portions of the ECG waveform, such as the QRS complex or ST segment, that contribute to anomalous predictions. 

Furthermore, AEs align well with TinyML requirements. They operate on fixed-size inputs and employ simple feedforward architectures that support parallel computation, resulting in predictable and low-latency inference. Their compact structure also minimizes memory usage and computational overhead during inference. 

The AE framework additionally provides architectural flexibility supporting both DNN and CNN based networks. A fully connected DNN-based AE efficiently captures global waveform characteristics, whereas a CNN-based AE can better capture localized spatial features within the ECG waveform. This flexibility enables trade-offs between model complexity, resource utilization, and detection performance based on application requirements.

During this work, alternative lightweight ML approaches, including Support Vector Machines (SVMs) and tree-based classifiers, were also considered. Although computationally efficient, such models rely on fixed decision boundaries learned from the training data~\cite{cortes1995support, breiman2001random}. As a result, they may fail to capture subtle waveform variations, particularly in single-beat ECG analysis where abnormal beats can closely resemble normal patterns. In contrast, the reconstruction-based formulation of AEs enables more robust identification of such deviations~\cite{hinton2006reducing, sakurada2014anomaly}.

Sequence-based models such as Long Short-Term Memory (LSTM) networks were also investigated. While effective in modeling temporal dependencies across multiple beats, their sequential processing nature introduces higher computational complexity and increased inference latency. These characteristics make them less suitable for low-power, real-time embedded deployment~\cite{hochreiter1997long,lane2015can,banbury2021benchmarking}.

Overall, AEs provide a balanced combination of anomaly detection capability, interpretability, architectural flexibility, and deployment efficiency. These characteristics make them well suited for beat-level ECG anomaly detection in embedded and wearable healthcare systems.
\subsection{Autoencoder Model Development and  Deployment}
\label{sec_sys_design_autoencoder_model}

In this work, two types of autoencoders were developed and evaluated for arrhythmia detection: a DNN-based autoencoder and a CNN-based autoencoder. In the DNN-based autoencoder, both the encoder and decoder blocks consist of fully connected (dense) neural network layers. In contrast, the CNN-based autoencoder employs convolutional layers in both the encoder and decoder blocks to better capture local temporal features in ECG signals. The detailed architectures of both models are presented in Section~\ref{sec_dnn_based_autoencoder}~and~\ref{sec_cnn_based_autoencoder}.

Both models were trained using normal ECG segments extracted from the MIT-BIH Arrhythmia Database. The objective was to learn a compact representation of normal ECG patterns by minimizing the reconstruction error during training.

After training, the models were evaluated on a dataset consisting of both normal and abnormal ECG segments from the MIT-BIH database. This evaluation was performed on the PC side using Python to verify anomaly detection performance prior to deployment. 

Following satisfactory floating-point performance, the trained models were prepared for hardware deployment. Deploying machine learning models on embedded systems requires additional optimization steps beyond standard model training due to the stringent constraints on memory, computational resources, and energy consumption. {This is particularly relevant for microcontroller-based systems, where available flash memory and RAM are typically limited to a few megabytes or less, floating-point computation support is often restricted, and are many cases powered from constrained energy sources.}

In this work, model optimization is primarily achieved through quantization, wherein model parameters are converted from 32-bit floating-point representations to lower-precision 8-bit integers. Quantization significantly reduces the memory footprint of the model and improves inference speed, both of which are critical requirements for microcontroller-based systems. Two common approaches to quantization are post-training quantization (PTQ) and quantization-aware training (QAT). Post-training quantization applies quantization after model training and may lead to noticeable degradation in model performance. In contrast, in QAT, quantization effects are simulated during the training process by inserting fake-quantization operations in the forward pass. This allows the model to adapt to reduced numerical precision and recover accuracy loss that typically occurs after quantization. As a result, QAT typically achieves overall better accuracy when compared to PTQ. This work therefore adopts QAT for optimizing the autoencoder model prior to deployment.

Post optimization, the models were deployed on an embedded platform. In this work, the ESP32-S3 microcontroller was selected as the target hardware platform. The trained models were converted to TensorFlow Lite format and subsequently integrated into the ESP32-S3 project framework by converting them into C arrays. Model inference was implemented using TensorFlow Lite for Microcontrollers (TFLite Micro), as provided within the Espressif ESP-IDF v5.4 framework via the ESP32 Arduino core. Further implementation details are provided in Section~\ref{sec_hardware_deployment}.

The deployed models were evaluated again, this time using ECG segments stored locally on the embedded platform. Performance metrics obtained from the embedded implementation were compared with the PC-side results to verify consistency between PC-side and on-device inference results.

In this work, hyperparameter tuning was performed manually to identify suitable model architectures for both CNN- and DNN-based autoencoders. A more systematic and comprehensive optimization of hyperparameters is left for future work. The architectures adopted in this study are described below.

\subsubsection{DNN-based Autoencoder} 
\label{sec_dnn_based_autoencoder}
The DNN-based autoencoder model consisted of encoder and decoder blocks, both constructed using fully connected neural network layers. During training, normal ECG segments from the dataset in Section~\ref{sec_dataset_training} were fed into the encoder, which compressed the signal into a lower-dimensional latent space. The encoder utilized three sequentially connected dense layers for this compression. The first layer contained 300 neurons, matching the length of the ECG segments in the dataset, while the second and third layers had 128 and 32 neurons, respectively. All neurons used ReLU activation. The compressed representation produced by the encoder was then fed into the decoder, which reconstructed the original ECG segment. The decoder was built using three sequentially connected dense layers, with 32, 128, and 300 neurons in the first, second, and third layers, respectively. Neurons in the first and second layers used ReLU activation, while those in the third layer used sigmoid activation. The model was trained to minimize the mean absolute error (MAE)~\cite{jadon2022comprehensive} between the normal ECG segments fed into the encoder and the reconstructed ECG segments at the output of the decoder.

\subsubsection{CNN-based Autoencoder} 
\label{sec_cnn_based_autoencoder}

The CNN-based autoencoder was designed to exploit local morphological patterns in ECG segments through convolutional feature extraction. Unlike the DNN-based autoencoder, which operates on flattened input vectors, the CNN-based model processes ECG segments in a structured two-dimensional representation to better capture local temporal correlations within the signal. Each ECG segment of length 300 samples was reshaped into a two-dimensional matrix of size $15 \times 20$ with a single channel, resulting in an input tensor of dimension $15 \times 20 \times 1$. This reshaping is a computational design choice that enables the application of two-dimensional convolutional filters while preserving local temporal continuity within the ECG signal. The encoder consisted of a sequence of convolutional and down-sampling layers. The first encoder layer applied 64 convolutional filters of size $3 \times 3$ with ReLU activation and same padding, followed by a $2 \times 2$ max-pooling layer to reduce spatial resolution. A second convolutional layer with 32 filters of size $3 \times 3$ and ReLU activation was then applied, followed by another $2 \times 2$ max-pooling layer. The encoder terminated with a convolutional layer comprising 16 filters of size $3 \times 3$, producing a compact feature representation that encoded salient ECG morphology. The decoder mirrored the encoder architecture to reconstruct the original ECG segment from the learned representation. It began with a convolutional layer containing 16 filters of size $3 \times 3$ with ReLU activation, followed by up-sampling layers to progressively restore the spatial resolution. A subsequent convolutional layer with 32 filters of size $3 \times 3$ and ReLU activation was applied, followed by another up-sampling layer. The final reconstruction was produced using a convolutional layer with a single filter of size $3 \times 3$ and sigmoid activation. Due to dimensional expansion introduced by up-sampling, a cropping operation was applied to obtain an output tensor of size $15 \times 20 \times 1$, which was then flattened to form a reconstructed ECG segment of length 300. Similar to the DNN-based autoencoder, the CNN-based autoencoder was trained exclusively using normal ECG segments from the dataset in Section~\ref{sec_dataset_training} to minimize the reconstruction error between the input and reconstructed ECG segments.

\section{Results}
\label{sec_results}

In this section, we discuss the implementation and evaluation details of the CNN-based and DNN-based autoencoder models for arrhythmia detection. We first describe the dataset engineering process, including ECG segmentation and extraction from the MIT-BIH database to construct the datasets used for model training and evaluation. This is followed by a discussion of model training and performance assessment across four experimental settings: (i) floating-point baseline evaluation on the PC side, (ii) INT8 quantized model evaluation using TensorFlow Lite simulation on the PC side, (iii) on-device inference on the ESP32-S3 microcontroller platform, and (iv) refined evaluation, where inference results are filtered to account for noise and variations in signal quality, ensuring a fair assessment under practical operating conditions.

\subsection{Dataset Engineering}
\label{sec_dataset_engineering}

\subsubsection{MIT-BIH Database}

The MIT-BIH Arrhythmia Database, hosted on the PhysioNet platform, is a widely used benchmark dataset for electrocardiogram (ECG) research, particularly for the development and evaluation of arrhythmia detection algorithms. The dataset was collected at the Arrhythmia Laboratory of Beth Israel Hospital, Boston, between 1975 and 1979 and consists of 48 half-hour excerpts of two-channel ambulatory ECG recordings acquired from 47 subjects and sampled at 360 Hz. Each record is accompanied by expert-marked R-peak locations and corresponding labels (annotations) identifying different beat types.

Among the various beat types labeled in the MIT-BIH Database, this work focuses on normal beats, \textit{N}, and a selected subset of abnormal beats, such as \textit{/}, \textit{L}, \textit{R}, \textit{A}, \textit{E}, and \textit{F}, for model training and evaluation. The beat label \textit{N} denotes normal sinus beats, \textit{/} represents paced beats, \textit{V} represents premature ventricular contractions, \textit{L} and \textit{R} correspond to left and right bundle branch block beats, \textit{A} denotes atrial premature beats, \textit{E} indicates ventricular escape beats, and \textit{F} represents fusion beats. These beat types were selected as they represent clinically relevant and frequently observed arrhythmic patterns with well-defined morphological characteristics. 

Beats labeled as non-cardiac events were excluded during dataset preparation, including pacemaker-related beats (\textit{f}), rhythm change or administrative markers (\textit{+}), non-beat or noise-related labels (\textit{\textasciitilde{}, !, ", x, |}), and segment-level rhythm labels (\textit{[} and \textit{]}). Rare beat types (\textit{a}, \textit{S}, \textit{e}) were also omitted to reduce class imbalance during model training. Premature ventricular contraction (PVC) beats labeled as \textit{V} were excluded due to inconsistencies observed between the assigned label positions and the R-peak locations.

\begin{figure*}[htbp]
  \centering
  \includegraphics[width=\linewidth]{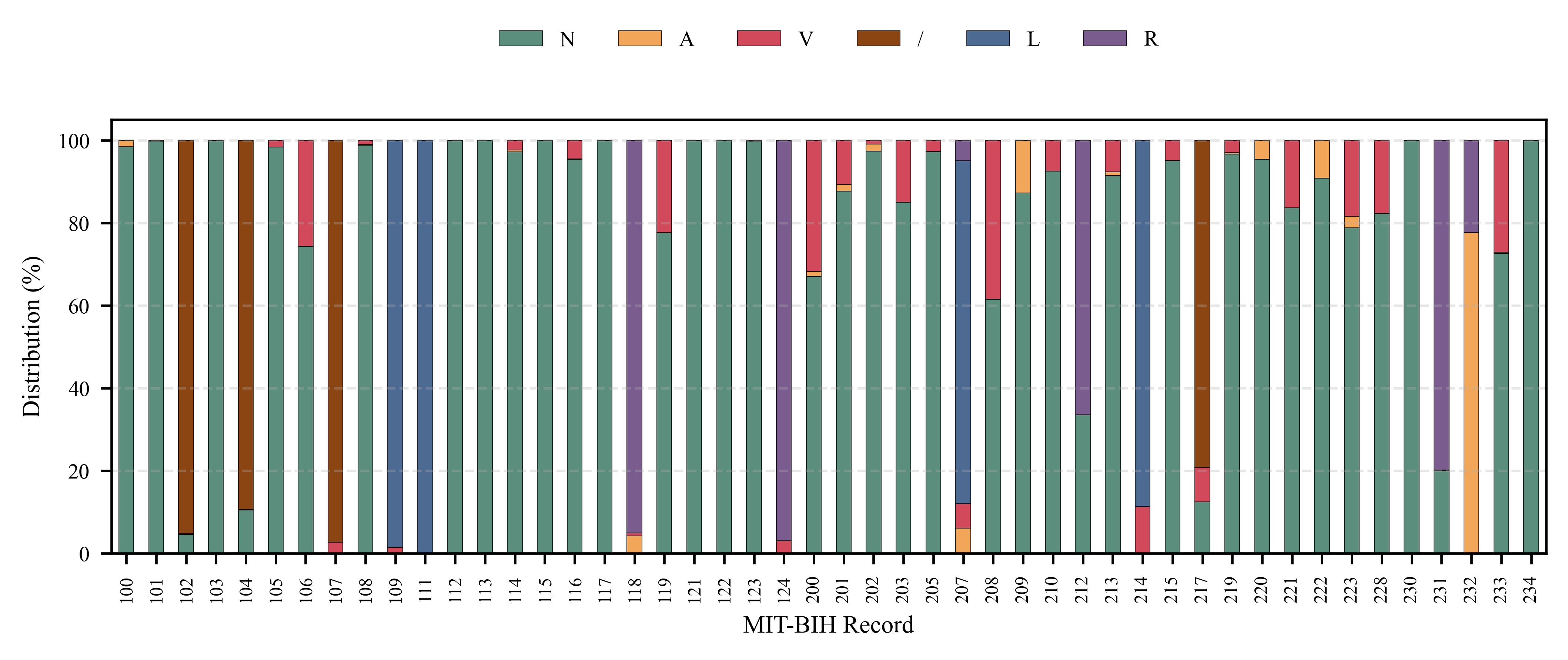}  
  \caption{Distribution of selected ECG beat types (\textit{N}, \textit{A}, \textit{V}, \textit{/}, \textit{L}, and \textit{R}) across individual patient records in the MIT-BIH Arrhythmia Database. The labels denote: N (normal sinus beats), A (atrial premature beats), V (premature ventricular contractions), / (paced beats), L (left bundle branch block beats), and R (right bundle branch block beats).}
  \label{fig_mit_statistics}
\end{figure*}

The distribution of these beats across patient records in the MIT-BIH Arrhythmia Database is illustrated in Fig.~\ref{fig_mit_statistics} and summarized in Table~\ref{tab:record_annotation_counts}. Fig.~\ref{fig_mit_statistics} presents the distribution of the selected beat types considered in this work (\textit{N}, \textit{/}, \textit{L}, \textit{R}, \textit{A}, \textit{E}, and \textit{F}), while Table~\ref{tab:record_annotation_counts} provides the distribution of all  annotated beat types in the dataset.

\begin{table*}[t]
\caption{Per-record annotation counts for different ECG beat types in the MIT-BIH Arrhythmia Dataset}
\label{tab:record_annotation_counts}
\centering
\scriptsize
\setlength{\tabcolsep}{3pt}
\renewcommand{\arraystretch}{0.95}

\begin{tabular}{rrrrrrrrrrrrrrrrrrrrrrrr}
\toprule
Record & ! & " & + & / & A & E & F & J & L & N & Q & R & S & V & [ & ] & a & e & f & j & x & \textbar & \textasciitilde \\
\midrule

100 & 0 & 0 & 1 & 0 & 33 & 0 & 0 & 0 & 0 & 2239 & 0 & 0 & 0 & 1 & 0 & 0 & 0 & 0 & 0 & 0 & 0 & 0 & 0 \\
101 & 0 & 0 & 1 & 0 & 3 & 0 & 0 & 0 & 0 & 1860 & 2 & 0 & 0 & 0 & 0 & 0 & 0 & 0 & 0 & 0 & 0 & 4 & 4 \\
102 & 0 & 0 & 5 & 2028 & 0 & 0 & 0 & 0 & 0 & 99 & 0 & 0 & 0 & 4 & 0 & 0 & 0 & 0 & 56 & 0 & 0 & 0 & 0 \\
103 & 0 & 0 & 1 & 0 & 2 & 0 & 0 & 0 & 0 & 2082 & 0 & 0 & 0 & 0 & 0 & 0 & 0 & 0 & 0 & 0 & 0 & 0 & 6 \\
104 & 0 & 0 & 45 & 1380 & 0 & 0 & 0 & 0 & 0 & 163 & 18 & 0 & 0 & 2 & 0 & 0 & 0 & 0 & 666 & 0 & 0 & 0 & 37 \\
105 & 0 & 0 & 1 & 0 & 0 & 0 & 0 & 0 & 0 & 2526 & 5 & 0 & 0 & 41 & 0 & 0 & 0 & 0 & 0 & 0 & 0 & 30 & 88 \\
106 & 0 & 0 & 41 & 0 & 0 & 0 & 0 & 0 & 0 & 1507 & 0 & 0 & 0 & 520 & 0 & 0 & 0 & 0 & 0 & 0 & 0 & 0 & 30 \\
107 & 0 & 0 & 1 & 2078 & 0 & 0 & 0 & 0 & 0 & 0 & 0 & 0 & 0 & 59 & 0 & 0 & 0 & 0 & 0 & 0 & 0 & 0 & 2 \\
108 & 0 & 0 & 1 & 0 & 4 & 0 & 2 & 0 & 0 & 1739 & 0 & 0 & 0 & 17 & 0 & 0 & 0 & 0 & 0 & 1 & 11 & 8 & 41 \\
109 & 0 & 0 & 1 & 0 & 0 & 0 & 2 & 0 & 2492 & 0 & 0 & 0 & 0 & 38 & 0 & 0 & 0 & 0 & 0 & 0 & 0 & 0 & 2 \\
111 & 0 & 0 & 1 & 0 & 0 & 0 & 0 & 0 & 2123 & 0 & 0 & 0 & 0 & 1 & 0 & 0 & 0 & 0 & 0 & 0 & 0 & 0 & 8 \\
112 & 0 & 0 & 1 & 0 & 2 & 0 & 0 & 0 & 0 & 2537 & 0 & 0 & 0 & 0 & 0 & 0 & 0 & 0 & 0 & 0 & 0 & 0 & 10 \\
113 & 0 & 0 & 1 & 0 & 0 & 0 & 0 & 0 & 0 & 1789 & 0 & 0 & 0 & 0 & 0 & 0 & 6 & 0 & 0 & 0 & 0 & 0 & 0 \\
114 & 0 & 0 & 3 & 0 & 10 & 0 & 4 & 2 & 0 & 1820 & 0 & 0 & 0 & 43 & 0 & 0 & 0 & 0 & 0 & 0 & 0 & 1 & 7 \\
115 & 0 & 0 & 1 & 0 & 0 & 0 & 0 & 0 & 0 & 1953 & 0 & 0 & 0 & 0 & 0 & 0 & 0 & 0 & 0 & 0 & 0 & 6 & 2 \\
116 & 0 & 0 & 1 & 0 & 1 & 0 & 0 & 0 & 0 & 2302 & 0 & 0 & 0 & 109 & 0 & 0 & 0 & 0 & 0 & 0 & 0 & 0 & 8 \\
117 & 0 & 0 & 1 & 0 & 1 & 0 & 0 & 0 & 0 & 1534 & 0 & 0 & 0 & 0 & 0 & 0 & 0 & 0 & 0 & 0 & 0 & 0 & 3 \\
118 & 0 & 0 & 1 & 0 & 96 & 0 & 0 & 0 & 0 & 0 & 0 & 2166 & 0 & 16 & 0 & 0 & 0 & 0 & 0 & 0 & 10 & 0 & 12 \\
119 & 0 & 0 & 103 & 0 & 0 & 0 & 0 & 0 & 0 & 1543 & 0 & 0 & 0 & 444 & 0 & 0 & 0 & 0 & 0 & 0 & 0 & 0 & 4 \\
121 & 0 & 0 & 1 & 0 & 1 & 0 & 0 & 0 & 0 & 1861 & 0 & 0 & 0 & 1 & 0 & 0 & 0 & 0 & 0 & 0 & 0 & 0 & 12 \\
122 & 0 & 0 & 1 & 0 & 0 & 0 & 0 & 0 & 0 & 2476 & 0 & 0 & 0 & 0 & 0 & 0 & 0 & 0 & 0 & 0 & 0 & 2 & 0 \\
123 & 0 & 0 & 1 & 0 & 0 & 0 & 0 & 0 & 0 & 1515 & 0 & 0 & 0 & 3 & 0 & 0 & 0 & 0 & 0 & 0 & 0 & 0 & 0 \\
124 & 0 & 0 & 13 & 0 & 2 & 0 & 5 & 29 & 0 & 0 & 0 & 1531 & 0 & 47 & 0 & 0 & 0 & 0 & 0 & 5 & 0 & 0 & 2 \\
200 & 0 & 0 & 148 & 0 & 30 & 0 & 2 & 0 & 0 & 1743 & 0 & 0 & 0 & 826 & 0 & 0 & 0 & 0 & 0 & 0 & 0 & 0 & 43 \\
201 & 0 & 0 & 35 & 0 & 30 & 0 & 2 & 1 & 0 & 1625 & 0 & 0 & 0 & 198 & 0 & 0 & 97 & 0 & 0 & 10 & 37 & 0 & 4 \\
202 & 0 & 0 & 8 & 0 & 36 & 0 & 1 & 0 & 0 & 2061 & 0 & 0 & 0 & 19 & 0 & 0 & 19 & 0 & 0 & 0 & 0 & 2 & 0 \\
203 & 0 & 0 & 45 & 0 & 0 & 0 & 1 & 0 & 0 & 2529 & 4 & 0 & 0 & 444 & 0 & 0 & 2 & 0 & 0 & 0 & 0 & 26 & 57 \\
205 & 0 & 0 & 13 & 0 & 3 & 0 & 11 & 0 & 0 & 2571 & 0 & 0 & 0 & 71 & 0 & 0 & 0 & 0 & 0 & 0 & 0 & 1 & 2 \\
207 & 472 & 0 & 24 & 0 & 107 & 105 & 0 & 0 & 1457 & 0 & 0 & 86 & 0 & 105 & 6 & 6 & 0 & 0 & 0 & 0 & 0 & 2 & 15 \\
208 & 0 & 0 & 53 & 0 & 0 & 0 & 373 & 0 & 0 & 1586 & 2 & 0 & 2 & 992 & 0 & 0 & 0 & 0 & 0 & 0 & 0 & 8 & 24 \\
209 & 0 & 0 & 21 & 0 & 383 & 0 & 0 & 0 & 0 & 2621 & 0 & 0 & 0 & 1 & 0 & 0 & 0 & 0 & 0 & 0 & 0 & 7 & 19 \\
210 & 0 & 0 & 17 & 0 & 0 & 1 & 10 & 0 & 0 & 2423 & 0 & 0 & 0 & 194 & 0 & 0 & 22 & 0 & 0 & 0 & 0 & 1 & 17 \\
212 & 0 & 0 & 1 & 0 & 0 & 0 & 0 & 0 & 0 & 923 & 0 & 1825 & 0 & 0 & 0 & 0 & 0 & 0 & 0 & 0 & 0 & 1 & 13 \\
213 & 0 & 0 & 43 & 0 & 25 & 0 & 362 & 0 & 0 & 2641 & 0 & 0 & 0 & 220 & 0 & 0 & 3 & 0 & 0 & 0 & 0 & 0 & 0 \\
214 & 0 & 1 & 25 & 0 & 0 & 0 & 1 & 0 & 2003 & 0 & 2 & 0 & 0 & 256 & 0 & 0 & 0 & 0 & 0 & 0 & 0 & 5 & 4 \\
215 & 0 & 2 & 5 & 0 & 3 & 0 & 1 & 0 & 0 & 3195 & 0 & 0 & 0 & 164 & 0 & 0 & 0 & 0 & 0 & 0 & 0 & 0 & 30 \\
217 & 0 & 0 & 67 & 1542 & 0 & 0 & 0 & 0 & 0 & 244 & 0 & 0 & 0 & 162 & 0 & 0 & 0 & 0 & 260 & 0 & 0 & 1 & 4 \\
219 & 0 & 4 & 21 & 0 & 7 & 0 & 1 & 0 & 0 & 2082 & 0 & 0 & 0 & 64 & 0 & 0 & 0 & 0 & 0 & 0 & 133 & 0 & 0 \\
220 & 0 & 0 & 17 & 0 & 94 & 0 & 0 & 0 & 0 & 1954 & 0 & 0 & 0 & 0 & 0 & 0 & 0 & 0 & 0 & 0 & 0 & 0 & 4 \\
221 & 0 & 0 & 23 & 0 & 0 & 0 & 0 & 0 & 0 & 2031 & 0 & 0 & 0 & 396 & 0 & 0 & 0 & 0 & 0 & 0 & 0 & 0 & 12 \\
222 & 0 & 0 & 136 & 0 & 208 & 0 & 0 & 1 & 0 & 2062 & 0 & 0 & 0 & 0 & 0 & 0 & 0 & 0 & 0 & 212 & 0 & 0 & 15 \\
223 & 0 & 0 & 28 & 0 & 72 & 0 & 14 & 0 & 0 & 2029 & 0 & 0 & 0 & 473 & 0 & 0 & 1 & 16 & 0 & 0 & 0 & 0 & 10 \\
228 & 0 & 3 & 41 & 0 & 3 & 0 & 0 & 0 & 0 & 1688 & 0 & 0 & 0 & 362 & 0 & 0 & 0 & 0 & 0 & 0 & 0 & 24 & 20 \\
230 & 0 & 0 & 207 & 0 & 0 & 0 & 0 & 0 & 0 & 2255 & 0 & 0 & 0 & 1 & 0 & 0 & 0 & 0 & 0 & 0 & 0 & 1 & 2 \\
231 & 0 & 427 & 11 & 0 & 1 & 0 & 0 & 0 & 0 & 314 & 0 & 1254 & 0 & 2 & 0 & 0 & 0 & 0 & 0 & 0 & 2 & 0 & 0 \\
232 & 0 & 0 & 1 & 0 & 1382 & 0 & 0 & 0 & 0 & 0 & 0 & 397 & 0 & 0 & 0 & 0 & 0 & 0 & 0 & 1 & 0 & 0 & 35 \\
233 & 0 & 0 & 71 & 0 & 7 & 0 & 11 & 0 & 0 & 2230 & 0 & 0 & 0 & 831 & 0 & 0 & 0 & 0 & 0 & 0 & 0 & 2 & 0 \\

234 & 0 & 0 & 3 & 0 & 0 & 0 & 0 & 50 & 0 & 2700 & 0 & 0 & 0 & 3 & 0 & 0 & 0 & 0 & 0 & 0 & 0 & 0 & 8 \\

\bottomrule
\\
\end{tabular}

\end{table*}

\paragraph*{Note} For classifying beats as normal, this work imposed an additional constraint whereby a minimum R–R interval of 170 samples (approximately 0.47 s at a sampling rate of 360 Hz) was enforced to limit the maximum heart rate to approximately 127 bpm. This constraint enforces physiologically plausible heart rates for normal sinus rhythm, reducing the likelihood of including ectopic or transient abnormal beats in the normal class~\cite{clifford2006advanced}. Clinically, normal sinus rhythm in adults is typically defined within the range of 60--100\,BPM ~\cite{goldberger2017clinical}.

\subsubsection{Dataset Preparation} 

In this work, two separate datasets were constructed: one for model training and another for model evaluation.
\paragraph{Training Dataset}
\label{sec_dataset_training}
For ML model training, a training dataset comprising normal and abnormal ECG segments was prepared using the MIT-BIH Arrhythmia Database. A Python-based preprocessing pipeline was developed for this purpose. The pipeline utilized PhysioNet’s Python libraries~\cite{Xie2023} to parse annotation files and identify R-peak locations and beat types.

Following R-peak identification, ECG segmentation was performed to extract cardiac cycles corresponding to normal and selected abnormal beat types. Conventionally, an ECG cycle consists of the P–QRS–T complex that captures the complete morphology of a heartbeat, including the P-wave, QRS complex, and T-wave. However, in this work, ECG segments were deliberately extracted using an R–R interval–based segmentation strategy.


In this work, ECG segmentation was performed as follows. If two consecutive R-peaks were both labeled as normal in the MIT-BIH database and the R–R interval contained at least 170 samples, the corresponding ECG segment was extracted and labeled as normal. If consecutive R-peaks were labeled as abnormal, the segment between them was extracted and labeled as abnormal. From each patient record, a maximum of 200 normal and 200 abnormal ECG segments were extracted to limit dataset imbalance across subjects. The final dataset consisted of 6,075 normal segments and 3,477 abnormal segments. The normal segments from this dataset were used to train the autoencoder-based models and the abnormal segments were used for validation during training to assess model performance.

The autoencoder-based ML model employed in this work requires input vectors of uniform length~\cite{goodfellow2016deep}. However, the extracted ECG segments varied in length due to natural variations in heart rate both within and across subjects. To address this, all ECG segments, normal and abnormal, were resampled to a fixed length of 300 samples. The resampled segments were aggregated into a single CSV file with binary labels to distinguish between normal and abnormal beats,  which served as the dataset for model training. In particular, the normal segments were used to train the autoencoder-based models and the abnormal segments were used for validation during training to assess model performance. Alongside the segments, additional metadata, including the patient record identifier, R-peak locations, and original segment length, were appended to facilitate traceability and subsequent failure analysis. Algorithm~\ref{alg:rr_pseudocode} details the procedure adopted for extracting normal ECG segments from a patient record. 

\begin{algorithm*}[!htbp]
\caption{R--R based beat extraction from ECG signals}
\label{alg:rr_pseudocode}
\renewcommand{\arraystretch}{1.15}
\begin{tabular}{|p{0.97\linewidth}|}
\hline
{\ttfamily
Input:\par
\ \ \ ecg $\rightarrow$ 1-D ECG signal (lead 0), sampled at 360 Hz\par
\ \ \ R[1..N] $\rightarrow$ list of R-peak sample indices from annotations\par
\ \ \ S[1..N] $\rightarrow$ list of beat symbols aligned to R-peaks\par
\ \ \ M $\rightarrow$ maximum number of segments per record\par
\ \ \ TARGET\_LEN $\rightarrow$ 300 (resampled length)\par
\par
Output:\par
\ \ \ Segments $\rightarrow$ list of standardized ECG segments\par
\par
Procedure:\par
1)\ Segments := empty\par
2)\ count := 0\par
3)\ For i = 1 to N--1:\par
\ \ \ - If S[i] = `N':\par
\ \ \ \ \ * require S[i+1] = `N' (two successive normals)\par
\ \ \ \ \ * require (R[i+1] -- R[i]) $\ge$ 170 samples\par
\ \ \ - start := R[i]\par
\ \ \ - end := R[i+1]\par
\ \ \ - seg := ecg[start:end]\par
\ \ \ - seg300 := linearly\_resample(seg, TARGET\_LEN)\par
\ \ \ - append seg300 to Segments\par
\ \ \ - count := count + 1;\ if count $\ge$ M: break\par
4)\ Return Segments\par
\par
Notes:\par
\ \ \ (+) Administrative marker; not a heartbeat → skipped.\par
\ \ \ (V) PVC beat; excluded during abnormal-beat extraction, since PVC annotations may not coincide with a consistent R-peak location, resulting in inaccurate R--R segment boundaries.\par
\ \ \ A minimum R--R interval of 170 samples ($\approx$0.47 s at 360 Hz) ensures that the normal beats correspond to heart rates below $\sim$127 bpm.\par
}\\
\hline
\end{tabular}
\end{algorithm*}

\paragraph{Evaluation Dataset}
\label{sec_dataset_evaluation}

For model evaluation, both in simulation and on hardware, a larger dataset of ECG segments from the MIT-BIH Arrhythmia Database was constructed. The dataset was prepared as follows. 

First, the ECG waveform from each patient record was bandpass filtered (Butterworth, 0.5–50 Hz) to suppress baseline wander and high-frequency noise. R-peaks were then identified, and were used to segment the ECG signal into R–R intervals. The R–R segments were then extracted and interpolated to a fixed length of 300 samples to ensure a uniform input size for the ML model.

The interpolated segments were stored as rows in a CSV file, along with beat labels extracted from the database corresponding to the left and right R-peaks defining each R–R segment. Additional metadata, including record identifiers and peak indices, were appended to each row to ensure traceability of the segments. The resulting dataset comprises over 95{,}000 ECG segments derived from 44 out of the 48 records in the MIT-BIH Arrhythmia Database.

During evaluation, each segment from the CSV file was classified by the trained ML model as either normal or abnormal. The predicted class labels were then compared against the ground-truth labels from the MIT-BIH database to compute performance metrics, including accuracy, precision, recall, and F1-score.

However, direct comparison between the model predictions and the ground truth requires careful consideration. In the MIT-BIH database  annotation records, labels are assigned to individual heartbeats corresponding to PQRST complexes around their R-peaks. In contrast, the proposed approach extracts R–R segments. Consequently, establishing correspondence between an R–R segment prediction and the labels requires alignment. Specifically, it must be determined whether the prediction for an R–R segment should be compared with the annotation associated with the left R-peak, the right R-peak, or a combination of both.

Three different comparison strategies have been investigated in our prior work~\cite{nagarajan2025autoencoder}. In the present study, we adopt the following comparison strategy, which performed better. The predicted classes for the left and right R–R segments are compared against the label corresponding to the R-peak. If the predicted classes for both adjacent R–R segments are normal and the annotation at the R-peak is normal, the classification is considered correct. If either of the adjacent R–R segments is predicted as abnormal and the annotation at the R-peak is abnormal, the classification is also considered correct. In all other cases, the prediction is considered incorrect, indicating a mismatch with the ground truth.

\begin{figure}[!htbp]
\centering
\includegraphics[width=\linewidth]{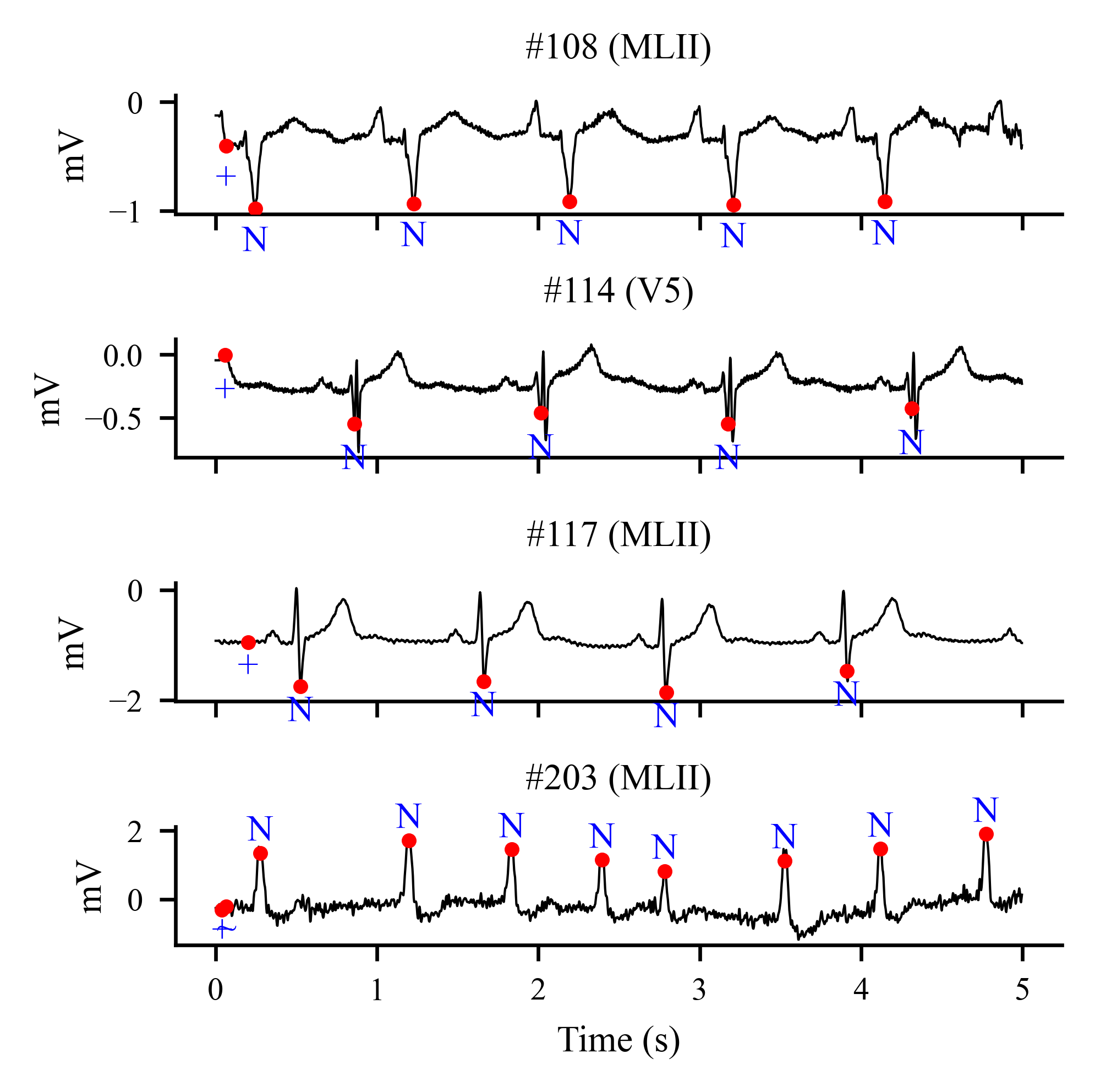}
\caption{Representative ECG segments from the omitted MIT-BIH records during dataset preparation, illustrating the reasons for exclusion: inverted QRS polarity in Record~\#108, lead configuration mismatch in Record~\#114, annotation inconsistency with shifted normal beat markers in Record~\#117, and significant high-frequency noise in Record~\#203.}
\label{fig:ecg_omitted_records}
\end{figure}

\paragraph*{Note} During dataset preparation, four patient records (\#108, \#114, \#117, and \#203) out of the 48 records in the MIT-BIH database were skipped from both the training and test datasets due to inconsistencies (see Fig.~\ref{fig:ecg_omitted_records}). Record \#108 was excluded because it contained an inverted QRS complex. Record \#114 was omitted due to a mismatch in ECG lead information (channel~1 corresponds to V5 instead of MLII), making it inconsistent with other records. Record \#117 had annotation inconsistencies; unlike other records, the normal annotations in Record \#117 appeared near the ECG ‘S’ peak instead of the ‘R’ peak. Finally, Record \#203 was excluded due to the presence of high-frequency noise in the ECG.

\paragraph*{Note} The evaluation dataset was derived from the MIT-BIH Database without relying on existing annotations, in order to better emulate real-time ECG processing scenarios where R-peak locations and class labels are not available a priori.

\subsection{Model Training and Evaluation}
\label{sec_model_training_and_evaluation}
This section discusses the training and performance evaluation of two autoencoder-based models for arrhythmia detection: a DNN-based autoencoder and a CNN-based autoencoder. It also presents the results obtained from the quantized variants of these models. 

\subsubsection{Baseline Models}
\label{sec_baseline_models}

The DNN- and CNN-based autoencoders discussed in Section~\ref{sec_sys_design_autoencoder_model} were trained in floating-point (FP32) precision using the normal ECG segments from the training dataset described in Section~\ref{sec_dataset_training}. Both models were optimized to minimize the reconstruction error between the input and reconstructed ECG segments. The Adam optimizer was used for training over 20 epochs with a batch size of 32. 

Post-training, both autoencoder models are expected to exhibit significant errors when processing abnormal ECG segments. To classify an ECG segment as normal or abnormal, it is thus sufficient to determine the reconstruction error. If the reconstruction error is significantly higher, the ECG segment is classified as abnormal; otherwise, it is classified as normal. The threshold for the reconstruction error, beyond which an ECG segment is declared abnormal, is determined as follows. First, the normal ECG segments used during training are passed through the model one by one, and the reconstruction error is measured for each pass. The threshold is then calculated as the mean plus one standard deviation of the errors.  

\begin{figure*}[!htbp]
    \centering
    \begin{subfigure}[b]{0.48\linewidth}
        \centering
        \includegraphics[width=\linewidth]{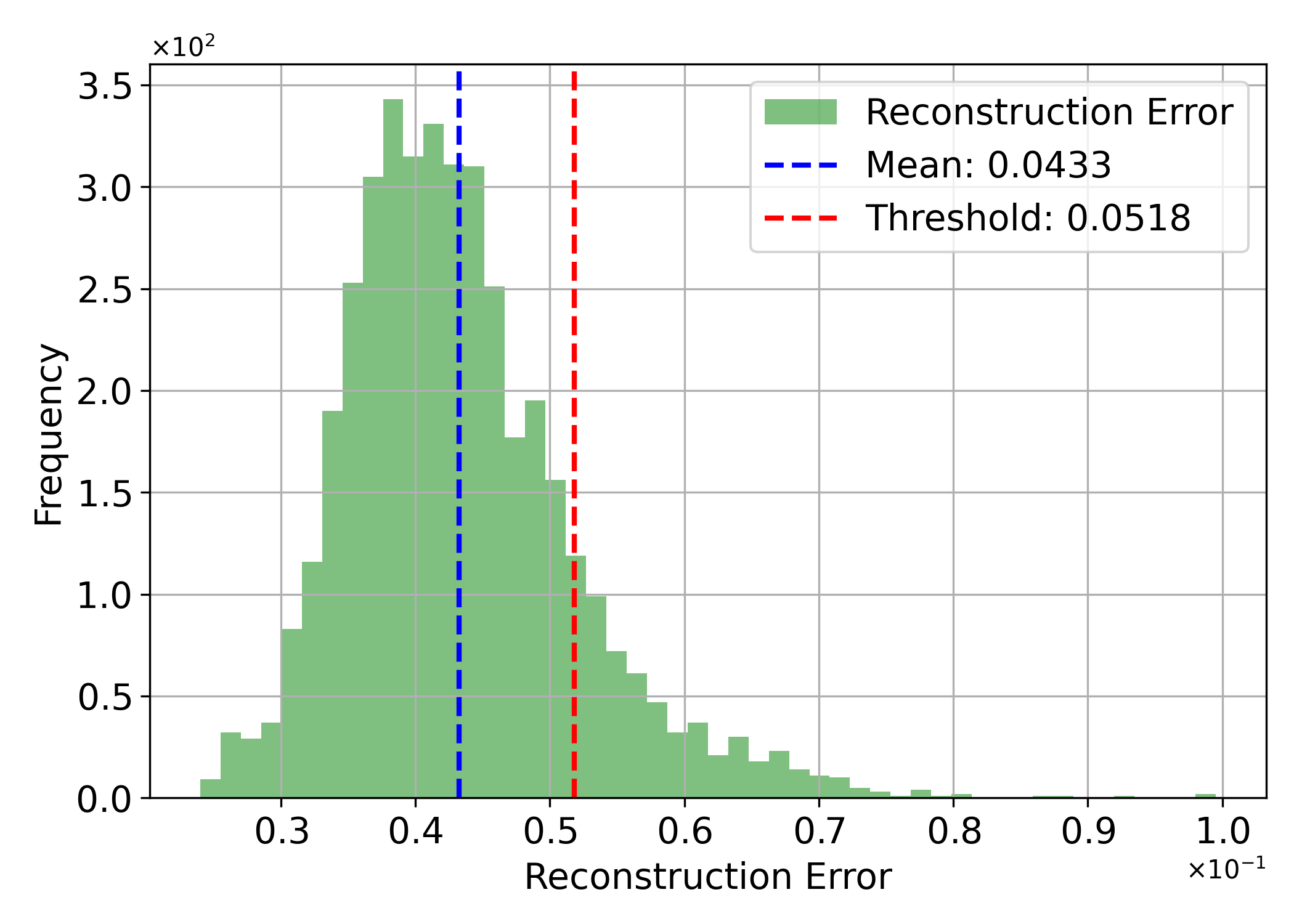}
        \caption{DNN-based autoencoder}
        \label{fig_reconstruction_error_hist_dnn}
    \end{subfigure}
    \hfill
    \begin{subfigure}[b]{0.48\linewidth}
        \centering
        \includegraphics[width=\linewidth]{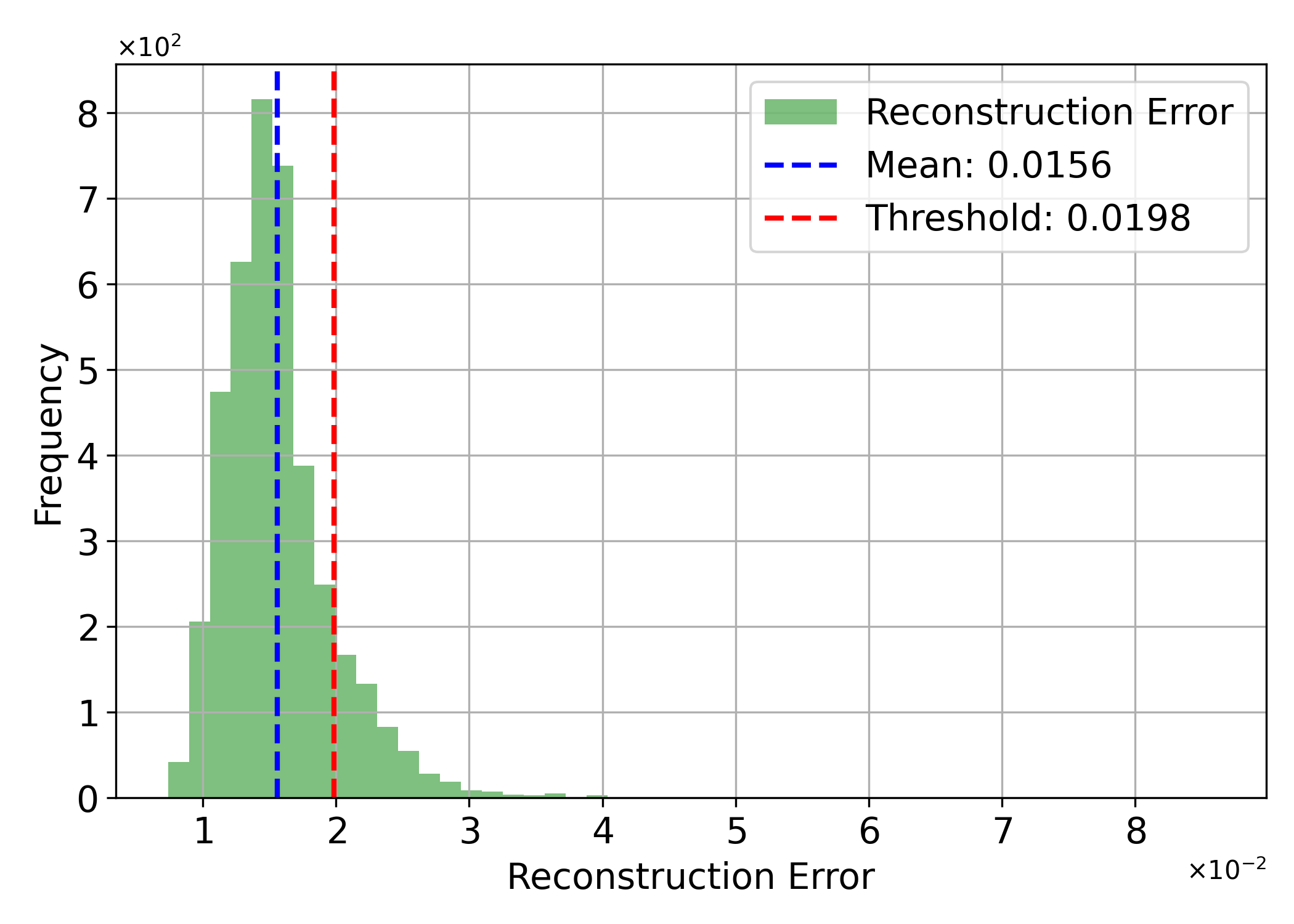}
        \caption{CNN-based autoencoder}
        \label{fig_reconstruction_error_hist_cnn}
    \end{subfigure}
    \caption{Distribution of reconstruction errors for normal ECG segments in the training dataset (Section~\ref{sec_dataset_engineering}) for (a) DNN-based and (b) CNN-based autoencoder models. The mean reconstruction error and the abnormal ECG segment detection threshold are indicated for both the models.}
    \label{fig_reconstruction_error_hist}
\end{figure*}

For evaluating the model’s performance in predicting abnormal ECG segments, we fed it a mix of normal and abnormal ECG segments from Section~\ref{sec_dataset_evaluation}. The reconstruction error threshold for classifying segments as normal or abnormal was set to 0.08192 for dnn-based and 0.01808 for cnn-based autoencoder model, respectively. A total of 1210 normal and 701 abnormal ECG segments, corresponding to 20\% of the training dataset, were used for the evaluation. Experiment~1 in Table~\ref{tab:perf_comparison} presents the results.

\subsubsection{Quantized Models}
\label{sec_quantized_models}

This work adopts QAT for optimizing the baseline autoencoder models discussed in Section~\ref{sec_baseline_models} prior to deployment. QAT for both DNN- and CNN-based autoencoders was implemented using the TensorFlow and Keras APIs via the TensorFlow Model Optimization Toolkit~\cite{tfmot}. The network layers were annotated for quantization to simulate INT8(8-bit) arithmetic during training, enabling the models to learn representations that are robust to reduced numerical precision. Both models were trained using the Adam optimizer with the objective of minimizing reconstruction loss for normal ECG segments. Post-training, the models were converted to TensorFlow Lite format and evaluated to simulate INT8 inference on a microcontroller-class device using the evaluation dataset described in Section \ref{sec_dataset_evaluation}.  Experiment~2 in Table~\ref{tab:perf_comparison} presents the results. Results indicate that quantization results in a fourfold reduction in model size and improvement in inference latency albeit with a potential trade-off in accuracy as expected due to lower-precision~\cite{jacob2018quantization, banbury2021micronets, tflite_micro}.

Fig.~\ref{fig_reconstruction_error_hist} shows the distribution of reconstruction errors for the normal ECG segments used in training for both the models, with the mean and thresholds marked.

\paragraph*{Note} During QAT, when the models were trained using normal segments from the training dataset described in Section~\ref{sec_dataset_training}, degraded performance was observed during evaluation. To address this issue, a filtered training dataset was used for QAT by removing atypical normal beats, reducing the number of normal segments from 6075 to 5188. For further details, refer to Section~\ref{sec_quantized_model_performance_analysis}.

\begin{figure}[!htbp]
    \centering
    \includegraphics[width=1\linewidth]{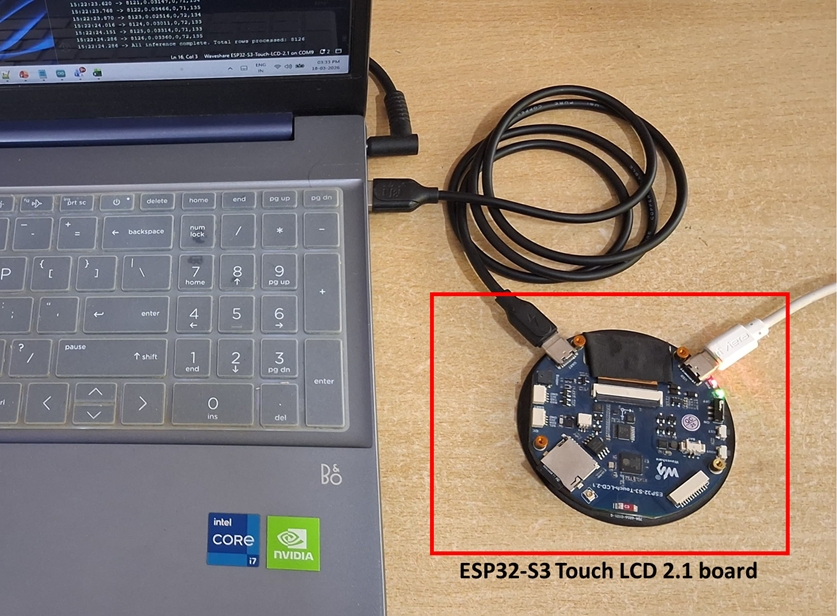}
    \caption{Hardware setup for evaluating on-device ECG inference using the ESP32-S3 Touch LCD 2.1 board with on-board Flash/PSRAM and microSD storage.}
    \label{fig_hw_setup}
\end{figure}

\begin{table*}[!htbp]
\centering
\caption{Comparison of performance metrics for different model variants evaluated in simulation and on embedded hardware.}
\label{tab:perf_comparison}
\renewcommand{\arraystretch}{1.2}
\begin{tabular}{|c|c|c|c|c|c|c|c|c|}
\hline
\textbf{Exp.} & \textbf{Platform} & \textbf{Model} & \textbf{Variant} 
& \textbf{Accuracy (\%)} & \textbf{Precision (\%)} & \textbf{Recall (\%)} & \textbf{F1-Score (\%)} & \textbf{Size (kB)} \\
\hline

\multirow{2}{*}{1}
& \multirow{2}{*}{PC-Side (Python Sim.)}
& DNN-based & Unquantized 
& 73.28 & 57.44 & 94.23 & 71.38 & 691 \\
\cline{3-9}
&
& CNN-based & Unquantized 
& 76.42 & 61.70 & 87.83 & 72.48 & 180 \\
\hline

\multirow{2}{*}{2}
& \multirow{2}{*}{PC-Side (Python Sim.)}
& DNN-based & Quantized 
& 69.79 & 54.66 & 85.16 & 66.59 & 180.1 \\
\cline{3-9}
&
& CNN-based & Quantized 
& 69.22 & 54.20 & 83.54 & 65.74 & 42.3 \\
\hline

\multirow{2}{*}{3}
& \multirow{2}{*}{Hardware (ESP32-S3)}
& DNN-based & Quantized 
& 69.68 & 54.55 & 85.36 & 66.56 & 180.1 \\
\cline{3-9}
&
& CNN-based & Quantized 
& 68.98 & 53.94 & 83.86 & 65.65 & 42.3 \\
\hline


\multirow{2}{*}{4}
& \multirow{2}{*}{Refined Evaluation}
& DNN-based & Quantized$^{\dagger}$ 
& 82.37 & 74.54 & 84.52 & 79.21 & -- \\
\cline{3-9}
&
& CNN-based & Quantized$^{\dagger}$ 
& 75.48 & 65.01 & 82.92 & 72.88 & -- \\
\hline

\end{tabular}

\vspace{0.3em}
\footnotesize{
$^{\dagger}$ Ambiguous segments excluded based on FP analysis. \\
}

\end{table*}

\subsection{Hardware Deployment and Evaluation}
\label{sec_hardware_deployment}
\subsubsection{Experimental Setup}

Quantized machine learning models from Section~\ref{sec_quantized_models} were deployed on hardware for evaluation. We selected the ESP32-S3 from Espressif Systems as the target microcontroller for deployment and evaluation of on-device inference. The ESP32-S3 is powered by a dual-core Xtensa\textsuperscript{\textregistered} LX7 32-bit RISC processor operating at up to 240~MHz and features 512~KB of on-chip SRAM, 16~MB of external Flash, and support for optional PSRAM~\cite{espressif_esp32s3}. 

Among the various development boards available for the ESP32-S3, we used the {Waveshare ESP32-S3 Touch LCD 2.1}~\cite{waveshare_esp32s3_lcd} development board due to its enhanced on-board memory support (16~MB of external Flash memory and 8~MB of external PSRAM), integrated LCD display, and built-in microSD card interface, features beneficial for data-driven inference workloads and subsequent debugging. Fig.~\ref{fig_hw_setup} shows the hardware setup used for the evaluation.

We used TensorFlow Lite for Microcontrollers (TFLite Micro) library ported for the ESP32 platform to deploy the quantized models and perform on-device inference. Deployment on the ESP32-S3 followed a standard TinyML workflow consisting of: (1) serialization of the trained and quantized model into a C-compatible byte array using the \texttt{xxd} utility; (2) integration of the serialized model into the ESP32 firmware as a static array stored in Flash memory; and (3) compilation and flashing of the firmware using the ESP-IDF toolchain, followed by runtime initialization of the inference engine and associated memory buffers. {ESP-32 firmware was written to support runtime selection and loading of both DNN- and CNN-based models, allowing flexibility based on evaluation needs. The firmware used approximately 744~KB of flash memory and 218~KB of RAM.}

{For evaluation, we used the evaluation dataset described in Section~\ref{sec_dataset_engineering}.} The dataset was stored on the development board’s on-board microSD card. Custom firmware was developed for the ESP32-S3 to sequentially load individual ECG segments from the SD card and perform inference using the deployed model.

During inference, the autoencoder reconstruction loss was computed for each ECG segment and compared against the predefined threshold in Fig.~\ref{fig_reconstruction_error_hist}. Based on this comparison, each segment was classified as either normal or arrhythmic.

\subsubsection{Evaluation Results}

{Exp.3 in Table~\ref{tab:perf_comparison} presents the results.} The results indicate that the DNN- and CNN-based autoencoder models achieved high recall values of 85\% and 83\% respectively. These findings show that the models, when executed directly on the hardware platform, are able to correctly identify the majority of abnormal ECG segments. The refined evaluation is discussed further in the Discussion section to reflect performance under practical operating conditions.

In terms of inference latency, on the ESP32-S3 operating at 240~MHz, the DNN model achieved an inference latency of 9 ms, while the CNN model required 71 ms per segment, with an additional overhead of approximately 68 ms due to SD card read/write operations.

\paragraph*{Note}
We also performed hardware-level validation of the autoencoder threshold used for classifying ECG segments as normal or abnormal. For this purpose, the training dataset described in Section~\ref{sec_dataset_engineering} was reintroduced to the model deployed on the hardware platform. Reconstruction losses were computed for each normal ECG segment, and the detection threshold was defined as the mean plus one standard deviation of these loss values. This threshold was then compared with the corresponding threshold obtained from PC-side inference. The thresholds obtained from PC-side and on-device inference were found to be in close agreement (DNN: 0.0518 vs. 0.0517; CNN: 0.0198 vs. 0.0197), confirming reliable deployment of the model on the hardware platform.

\subsection{Refined Evaluation Post Failure Analysis}
\label{sec_refined_evaluation}

Post deployment on hardware, evaluation of the model exhibited good recall; however, the accuracy and precision were lower than expected. This degradation was primarily due to an increased number of false predictions (false positives and false negatives). To better understand these outcomes, a detailed record-wise failure analysis was conducted to characterize model behavior across diverse ECG morphologies and rhythm patterns.

The analysis revealed several key observations. In multiple cases, apparent misclassifications corresponded to early or subtle arrhythmic patterns that were labeled as normal in the reference annotations. Although these instances contributed to higher false positives, the model’s ability to flag such patterns indicates sensitivity to early-stage arrhythmic conditions.

Additionally, since the proposed system relies on single-lead ECG acquisition to emulate real-world wearable scenarios, certain abnormal conditions that appear indistinguishable from normal rhythms in a single lead were incorrectly classified as normal, contributing to false negatives. This limitation arises from the ECG acquisition configuration rather than the model itself.

Furthermore, a small number of incorrect predictions were attributed to noisy ECG waveforms and inconsistencies in dataset annotations. A comprehensive discussion of these observations is provided in Section~\ref{sec_discussion}.

To obtain fairer and more representative performance metrics, the evaluation was repeated after omitting certain records from the evaluation dataset. Records affected by significant noise and labeling inconsistencies were removed, as described below.

Based on the false positive analysis, records associated with morphology-driven and signal-induced effects (Table~\ref{tab:fp_analysis} (b) and (c)) were excluded. In these records, waveform distortions, baseline wander, and atypical morphologies biased the reconstruction error, leading to inflated false positives.

Ectopic-driven cases (Table~\ref{tab:fp_analysis} (a)) were handled separately by accounting for segmentation effects. Specifically, ventricular beats (\texttt{V}) were addressed by excluding neighboring normal (\texttt{N}) annotations to avoid mixed-beat segments arising from closely spaced R--R intervals.

In addition, administrative markers and non-beat annotations (e.g., \texttt{[" , \textasciitilde , +, [, ], |, !]}) were excluded, as they correspond to signal quality indicators, rhythm transitions, or annotation artifacts rather than true cardiac activity.

Exp.4 in Table~\ref{tab:perf_comparison} presents the results of this refined evaluation. The results indicate that the DNN-based autoencoder model demonstrates strong recall, precision, and accuracy values of 84\%, 74\%, and 82\%, respectively.

\begin{figure}[!htbp]
    \centering
    \includegraphics[width=1\linewidth]{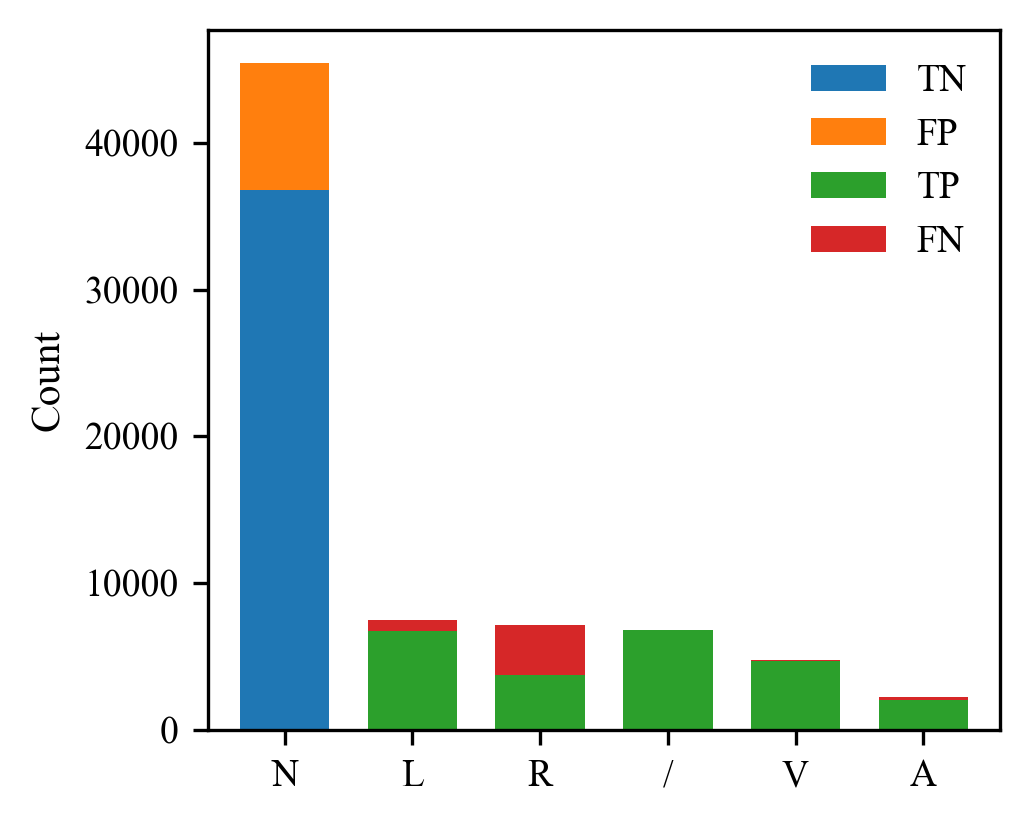}
\caption{Annotation-level classification counts of true positives (TP), true negatives (TN), false positives (FP), and false negatives (FN) obtained from the refined evaluation. Only arrhythmia types with sufficient sample counts are included.}    \label{fig:annotation_error_distribution}
\end{figure}

Fig.~\ref{fig:annotation_error_distribution} presents the annotation-level counts of true positives (TP), true negatives (TN), false positives (FP), and false negatives (FN) obtained after refined evaluation.

\paragraph*{Note} For the refined evaluation, we did not omit beat types that are not reliably represented under single-lead, single-beat constraints. In particular, bundle branch block variants (\texttt{L, R}) are difficult to distinguish using MLII morphology alone, and atrial premature beats (\texttt{A}) depend on R--R interval variability that is not captured in isolated beat-based analysis, we did not omit these from the evaluation records in order to capture the realistic system performance metrics expected from MLII-based wearable systems.

\begin{table*}[!htbp]
\centering
\caption{Analysis of False Positives (FP): Ectopic-Beat--Driven, Morphology-Driven, and Signal-Induced Effects}
\label{tab:fp_analysis}
\begin{tabular}{c c c c c p{8.5cm}}
\hline
\textbf{Record} & \textbf{FP} & \textbf{TP} & \textbf{V} & \textbf{N} & \textbf{Description} \\
\hline

\multicolumn{6}{l}{\textbf{(a) Ectopic Beat--Driven (PVC-driven) False Positives:} High ventricular beat activity causing deviation from normal ECG morphology} \\
\#200 & 1724 & 1030 & 826 & 1743 & R--R--based segmentation may merge adjacent normal and ectopic beats into a single segment, increasing false positives. \\
\#208 & 1492 & 1443 & 992 & 1586 & Same as above \\
\#233 & 1401 & 914  & 831 & 2230 & Same as above \\
\#106 & 899  & 569  & 520 & 1507 & Same as above \\

\hline
\multicolumn{6}{l}{\textbf{(b) Morphology-Driven False Positives:} Waveform morphology deviations influencing model output} \\
\#222 & 1694 & 472 & 0   & 2062 & Prominent P-wave morphology suggesting atrial variation/enlargement \\
\#228 & 1681 & 449 & 362 & 1688 & ST-elevation--like morphology; possible early repolarization, clinical correlation required \\
\#105 & 1224 & 129 & 41  & 2526 & ST-segment sagging with altered T-wave morphology \\
\#202 & 819  & 79  & 19  & 2061 & Subtle morphology variation (slightly reduced PR interval) \\

\hline
\multicolumn{6}{l}{\textbf{(c) Signal / Biasing / Distortion-Induced False Positives:} Signal quality effects contributing to FP} \\
\#213 & 2484 & 642 & 220 & 2641 & Abnormal ECG recording with significant waveform distortion (baseline/filtering effects); ST depression noted; \\
\#122 & 1740 & 2   & 0   & 2476 & Tall P-wave observed; further evaluation required, correlation with echocardiographic findings recommended. Near-normal morphology \\
\#121 & 1316 & 13  & 1   & 1861 & Near-normal ECG; false positives likely due to baseline wander \\

\hline
\end{tabular}
\par\vspace{2pt}
\noindent{\footnotesize FP and TP denote model predictions w.r.t. MIT-BIH annotations; 
V and N indicate ventricular and normal beat counts.}
\end{table*}

\begin{table*}[!htbp]
\centering
\caption{Analysis of False Negatives (FN): Morphology Similarity and Lead/MLII Effects}

\label{tab:fn_analysis}
\begin{tabular}{c c c c c c c c c l}
\hline
\textbf{Record} & \textbf{FN} & \textbf{TP} & \textbf{TN} & \textbf{FP} & \textbf{L} & \textbf{R} & \textbf{V} & \textbf{N} & \textbf{Description} \\
\hline

\#124 & 1432 & 202  & 0   & 0   & 0    & 1531 & 47  & 0    & Narrow QRS; R-dominant, MLII lead resembles normal morphology \\

\#231 & 1275 & 422  & 2   & 312 & 0    & 1254 & 2   & 314  & Mixed narrow QRS; R/N variability, morphology overlaps in MLII and V1 \\

\#214 & 844  & 1453 & 0   & 0   & 2003 & 0    & 256 & 0    & Wide QRS; L-dominant, ventricular morphology prominent in V1 \\

\#118 & 757  & 1544 & 0   & 0   & 0    & 2166 & 16  & 0    & Narrow QRS; strong R-dominance, MLII resembles normal beats \\

\hline
\end{tabular}
{\footnotesize FN, TP, TN, and FP are model predictions w.r.t. MIT-BIH annotations; \\
L, R, V, and N denote left/right bundle branch, ventricular, and normal beat counts.}
\end{table*}

\begin{figure*}[!htbp]
	\centering
	\includegraphics[width=\textwidth]{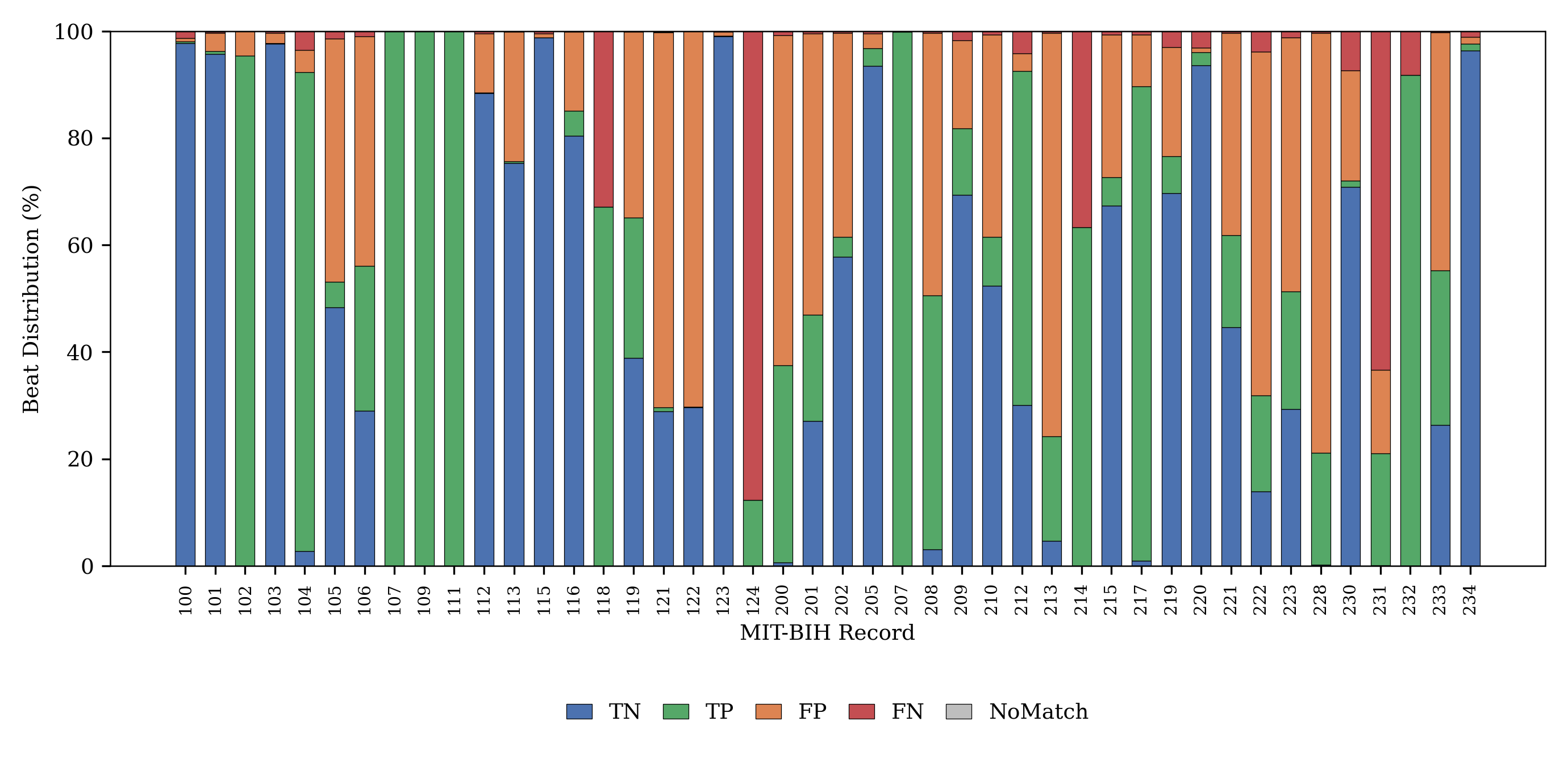}
	\caption{Record-wise distribution of true positives, true negatives, false positives, and false negatives across the evaluated ECG recordings for the DNN-based autoencoder model.} 
	\label{fig_fa_chart}
\end{figure*}

\section{Discussion}
\label{sec_discussion}

\label{sec_post_deployment_analysis}
This section analyzes quantized model behavior under practical deployment conditions, using real-world ECG data from the MIT-BIH dataset, with emphasis on segment-level interpretation and record-specific variations.

Several practical insights were gained during the development and evaluation of the proposed on-device ECG anomaly detection framework. The experiments show that quantized model behavior is influenced by record-specific characteristics such as rhythm composition, beat morphology diversity, and signal quality. These observations highlight the importance of record-wise analysis of model performance.

\subsection{Quantized Model Performance Analysis}
\label{sec_quantized_model_performance_analysis}

The proposed unquantized autoencoder model (i.e., the original floating-point model) was intentionally compact (e.g., fewer than 1{,}000 neurons in the DNN configuration) to meet TinyML deployment constraints. While this enables efficient on-device inference, it also makes the model more sensitive to quantization effects when converting from floating-point (FP32) to reduced-precision (INT8) representation.

Using the dataset in Section~\ref{sec_dataset_training} with quantized models resulted in a drop in key performance metrics, including recall. Due to the limited model capacity, quantization-induced precision loss affects the reconstruction of subtle waveform variations.

To address this, prediction errors were systematically analyzed and the model was refined using an improved training dataset. Normal segments in arrhythmia-rich records were not included to avoid ambiguous or atypical morphology during training (particularly in PVC-dominant records). In addition, segments with atypical QRS width were excluded to reduce variability arising from morphology that is not consistently represented in the dataset. These refinements improved model performance under reduced precision.

The deployment of the autoencoder model in a quantized TinyML configuration introduced additional challenges during evaluation on the MIT-BIH Arrhythmia Database. These are primarily driven by variations in ECG morphology and rhythm patterns. Variations in QRS complex width, such as unusually wide or narrow morphologies, were sometimes reconstructed accurately by the autoencoder, resulting in increased false negative detections. Similarly, a small subset of paced beats, which often resemble normal QRS morphology, contributed to misclassifications due to limited morphological distinction in single-lead representation.

Additional challenges were observed in records exhibiting repeating ventricular rhythm patterns such as bigeminy and trigeminy (e.g., N--V--N--V or N--N--V sequences). These patterns introduce non-typical waveform distributions. This behavior is influenced by the R--R segment-based analysis, where a single segment may contain more than one beat types (e.g., normal and ventricular). In the current binary detection formulation, such segments are treated as arrhythmic, which remains consistent with the overall detection objective. However, this limits the ability to distinguish individual beat types within a segment. Addressing this limitation through finer segmentation or multiclass autoencoder formulations remains part of ongoing work.

Overall, these observations show that quantization affects the model’s ability to capture subtle ECG morphology, and that careful dataset refinement is necessary to maintain reliable performance in embedded deployment.

\subsection{Failure Case Analysis}

Post deployment on hardware in Section~\ref{sec_quantized_models}, the model exhibited good recall; however, precision and overall accuracy were lower than expected due to an increased number of false positive and false negative detections. This behavior was rectified after omitting certain records used for evaluating the model as discussed in Section~\ref{sec_refined_evaluation}. The records to omit were identified through a detailed failure case analysis. This section discusses this analysis in detail. 

For helping in the failure analysis, several analyses were performed, including inspection of record-wise confusion matrix metrics, inspection of stacked bar chart visualizations of classification outcomes, and statistical analysis of record-wise mean QRS widths. In addition, records contributing the highest false negative and false positive counts were examined to identify those contributing disproportionately to classification errors. These analyses provide insight into how morphology and rhythm patterns influence reconstruction-based anomaly detection performance in quantized embedded models. 

Fig.~\ref{fig_fa_chart} presents the record-wise stacked bar chart. Each bar corresponds to a single ECG record, and the total height of the bar represents the complete set of evaluated beats for that record. Each colored segment within a bar denotes the proportion of beats classified as true positives (TP), true negatives (TN), false positives (FP), false negatives (FN), and NoMatch. By visualizing these categories together, the chart provides an integrated view of both correct classifications and error modes on a per-record basis, enabling direct comparison of performance characteristics across different ECG recordings.

The chart shows that several records are dominated by TN outcomes, indicating reliable identification of normal beats. In contrast, records with a higher proportion of abnormal rhythms demonstrate increased TP counts, confirming the model’s ability to detect anomalous beats. However, certain records also exhibit elevated FP or FN proportions, reflecting challenges in distinguishing subtle morphological variations from true anomalies. These misclassifications were primarily associated with records showing waveform variability, abrupt morphological transitions, or irregular rhythm patterns. This observation reinforces the need for cautious interpretation of beat-level results in continuous ECG analysis.

To further investigate the causes of these misclassifications, selected ECG segments from representative records were examined in detail  to analyze false positive and false negative detections.

\subsubsection{False Positive Cases} 
\begin{figure*}[!t]
\centering

\begin{subfigure}{0.48\linewidth}
    \includegraphics[width=\linewidth]{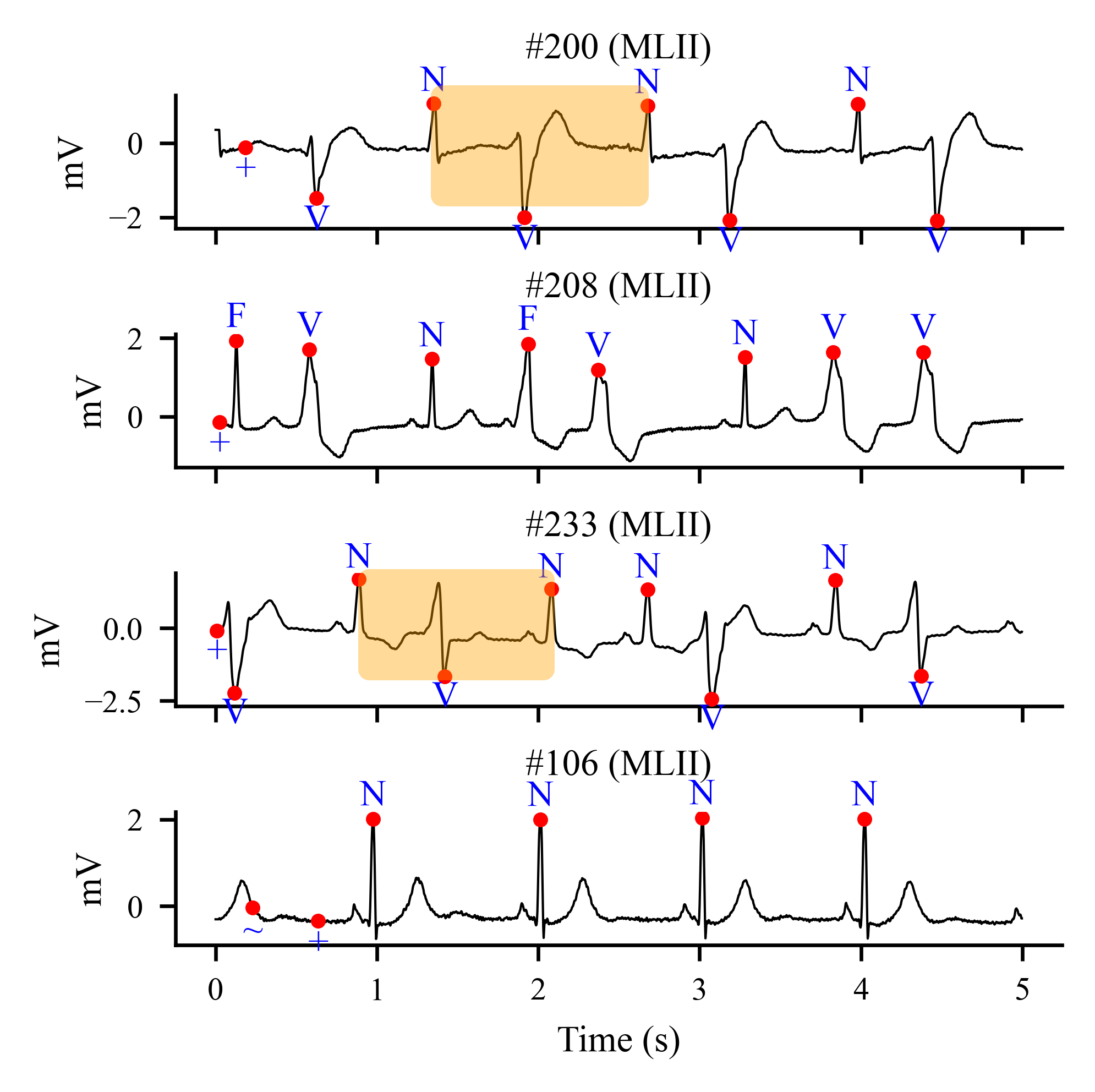}
    \caption{Ectopic beat–driven (PVC-driven) cases: R--R segments containing both normal (N) and ventricular (V) beats within the same interval, leading to combined N and V activity representative of arrhythmic patterns. Refer to the discussion section for more details.}
\end{subfigure}
\hfill
\begin{subfigure}{0.48\linewidth}
    \includegraphics[width=\linewidth]{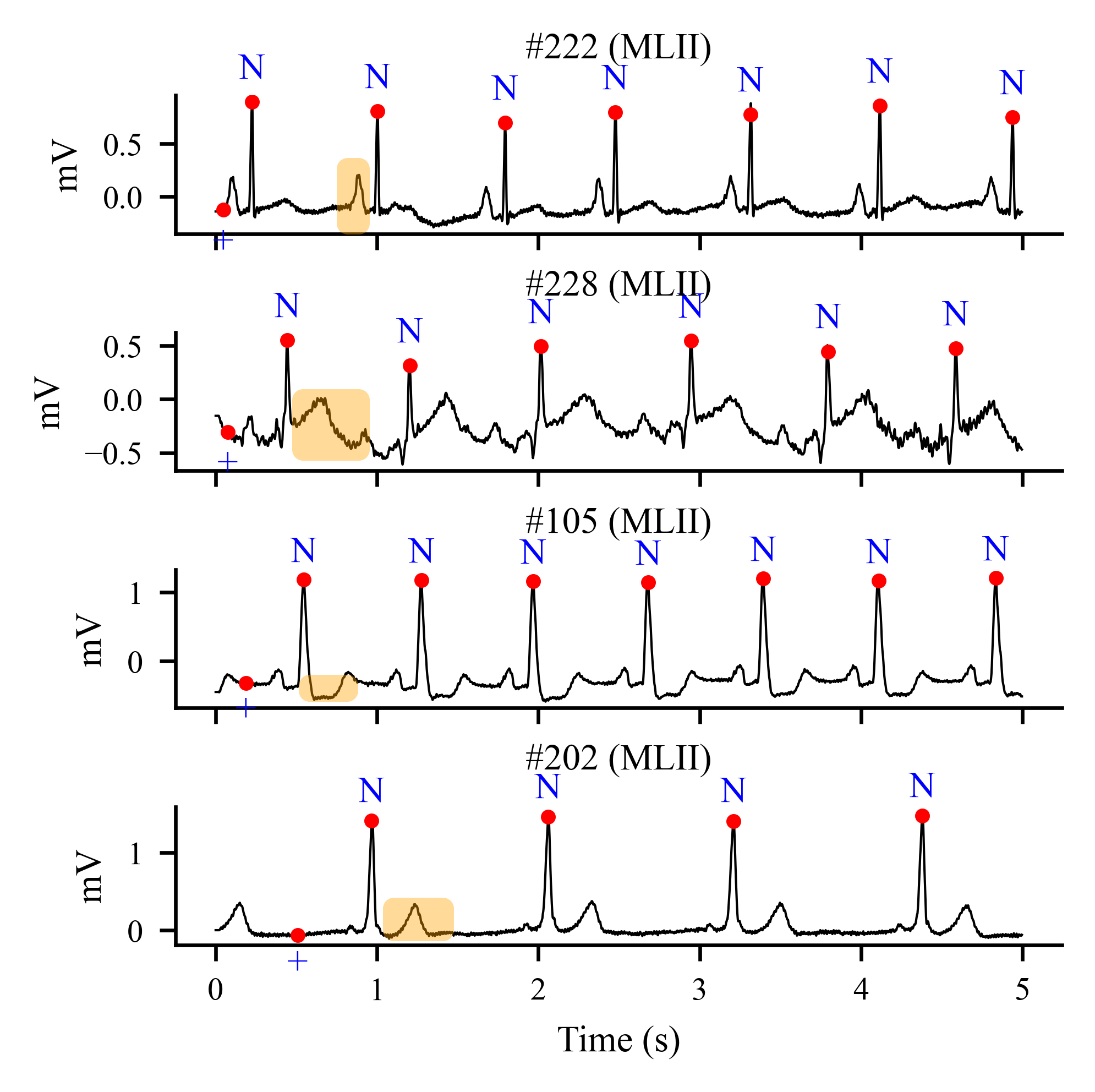}
    \caption{Morphology-driven false positives (FP): prominent P-wave variation (Record~\#222), ST-elevation–like morphology (Record~\#228), ST-segment sagging with T-wave changes (Record~\#105), and subtle morphology variation (Record~\#202).}
\end{subfigure}


\begin{subfigure}{0.48\linewidth}
    \includegraphics[width=\linewidth]{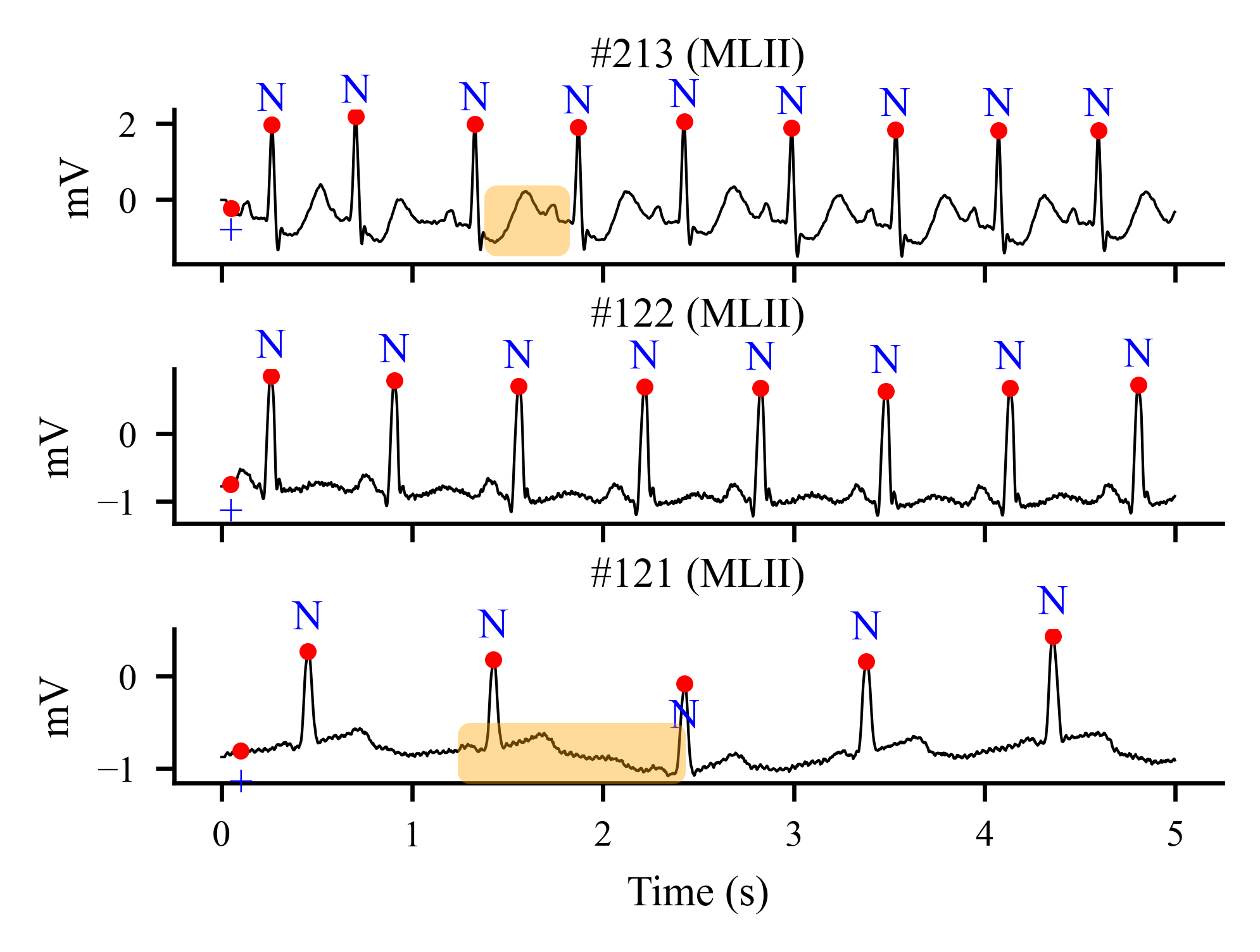}
    \caption{Signal-induced false positives (FP): waveform distortion (Record~\#213), Tall P-wave observed and near-normal patterns (Record~\#122), baseline wander (Record~\#121) and normal rhythm reference (Record~\#115).}
\end{subfigure}
\hfill
\begin{subfigure}{0.48\linewidth}
    \includegraphics[width=\linewidth]{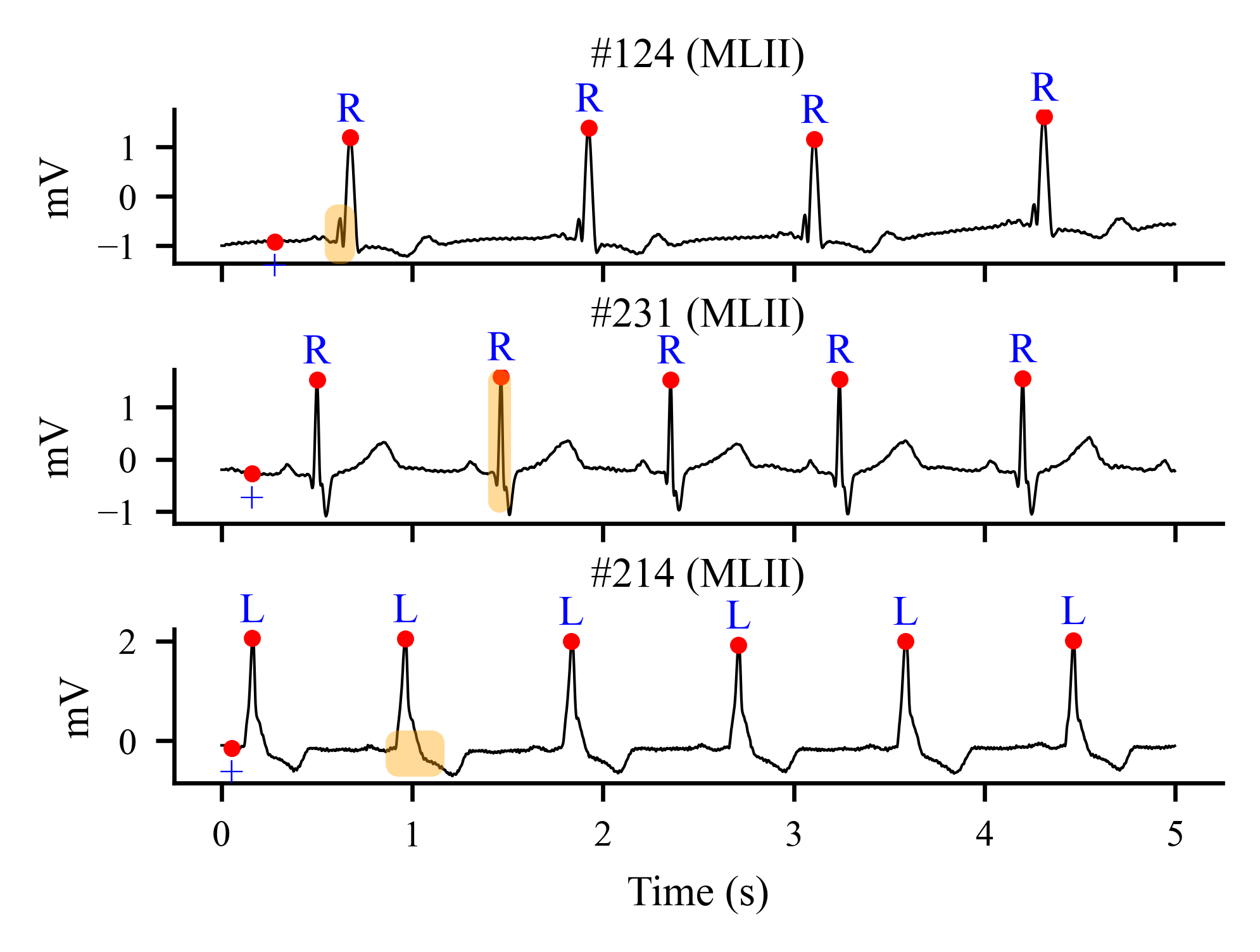}
    \caption{False negative (FN) cases in the MLII lead, including R-dominant patterns (Records~\#124) where the autoencoder shows reduced sensitivity to sharp, short pre-QRS spikes, mixed R/N patterns (Record~\#231), and L-dominant ventricular patterns (Record~\#214).}
\end{subfigure}
\caption{Illustrative ECG examples highlighting false positive (FP) and false negative (FN) cases across selected MIT-BIH records, with corresponding MIT-BIH annotation labels indicated in each plot.}
\label{fig:ecg_fp_fn}
\end{figure*}

False positive detections were predominantly observed in records containing rhythm irregularities or morphological variations that deviate from typical sinus rhythm patterns. In such cases, the autoencoder reconstruction error increases even though the reference annotation labels the beat as normal. Table~\ref{tab:fp_analysis} (a)--(c) summarizes the false positive (FP) cases, while Fig.~\ref{fig:ecg_fp_fn}(a)--(c) illustrates representative ECG segments corresponding to these cases, with annotation labels indicated in each plot.

For example, Records~\#200 and \#208 contain frequent ventricular ectopy forming repeating rhythm patterns such as bigeminy and trigeminy (e.g., N--V--N--V or N--N--V sequences, see Fig.~\ref{fig:ecg_fp_fn}(a)). In such rhythms, segmentation based on consecutive R--R intervals may capture segments with mixed beat compositions, resulting in variations in the waveform within a single segment. These patterns, however, remain indicative of underlying arrhythmic behavior at the rhythm level.

Record~\#222 contains ECG waveforms with tall and prominent P-waves suggesting potential atrial morphological variations. Similarly, Record~\#228 contains segments with mild ST-segment deviations that could resemble early ischemic patterns. For records such as Record~\#222 and Record~\#228, although these patterns may not always be labeled as abnormal, they differ from the learned normal morphology, resulting in false positive detections (see Fig.~\ref{fig:ecg_fp_fn}(b)). Such waveform variations may also arise from benign physiological conditions or medication-related effects and therefore require clinical correlation for interpretation. These cases indicate that the model is sensitive to subtle waveform deviations that may precede significant abnormalities.

Fig.~\ref{fig:ecg_fp_fn}(c) presents signal- and distortion-induced false positive cases influenced by waveform quality effects. Record~\#213 shows an abnormal ECG with significant waveform distortion, including baseline and filtering effects, along with ST-segment depression. Record~\#122 exhibits near-normal morphology with a prominent P-wave, where subtle morphological variations contribute to misclassification. Similarly, Record~\#121 represents a near-normal ECG affected by baseline wander, leading to deviations in the waveform. These cases highlight how signal quality and subtle waveform variations can influence classification outcomes and contribute to false positive detections.

\subsubsection{False Negative Cases} False negative detections were primarily observed in records where abnormal beats retain waveform structures similar to normal sinus beats. In such situations, the autoencoder is able to reconstruct the abnormal waveform with relatively low error, causing the anomaly score to fall below the detection threshold. Fig.~\ref{fig:ecg_fp_fn}(d) shows a representative ECG plot corresponding to a false negative (FN) case, with corresponding MIT-BIH annotation labels indicated.

Record~\#124 represents one such case, where R-beats appear morphologically similar to normal QRS complexes in the MLII lead. 

Record~\#231 contains a mixture of normal sinus beats and ventricular ectopic beats, with several abnormal beats exhibiting QRS complexes that are only moderately widened and retain morphological characteristics similar to normal sinus QRS complexes. This similarity reduces reconstruction error and increases the likelihood of false negative detections in reconstruction-based anomaly detection approaches.

In Records~\#214 and \#118, most abnormal segments are correctly detected, but a subset is reconstructed with low error due to morphological similarity, leading to occasional missed detections.

Bundle Branch Block patterns (\texttt{L}, \texttt{R}):
The left and right bundle branch block patterns are challenging to distinguish when using the MLII lead alone. While QRS widening (e.g., $>$120\,ms) is commonly associated with complete bundle branch block, incomplete bundle branch block patterns may exhibit QRS durations below this threshold~\cite{aha_ecg_standards, goldberger2017clinical}. Accurate differentiation therefore requires multi-lead information (e.g., V1, V6) beyond QRS duration alone. Right bundle branch block (RBBB) patterns are most prominently observed in lead V1, whereas left bundle branch block (LBBB) produces broader waveform changes visible across multiple leads~\cite{wagner_ecg}. As a result, certain morphology-specific variations, particularly RBBB, may lead to missed detections.

\paragraph*{Note} It is noted that the MIT-BIH Arrhythmia Database has undergone annotation corrections and refinements a few times over its release history, including updates to beat labels and rhythm annotations across several records (e.g., records 119, 203, 214, and 222), as documented in the PhysioNet revision history~\cite{Moody2001}. In particular, in Record~\#214, left bundle branch block beats were originally labeled as normal and later corrected, as reported in the PhysioNet documentation~\cite{Moody2001}. In this work, we use the current PhysioNet release (v1.0), which incorporates these corrections.

\subsubsection{Additional Annotation-Specific Observations}

While the majority of false predictions are driven by morphology and rhythm variations, a smaller subset of errors is associated with specific annotation types in the MIT-BIH dataset. These cases are limited in number but provide useful insights into model behavior.

Pacemaker Spikes (\texttt{"}):
This annotation denotes narrow pacemaker spikes preceding the QRS complex. The model shows limited sensitivity to these sharp, low-amplitude artifacts and instead relies primarily on overall waveform morphology. As a result, a small portion of segments containing subtle pacing spikes are classified as normal. Approximately 400+ such annotations are observed, with representative occurrences in records such as \#102, \#104, and \#217.

Paced Beats (\texttt{/}):
Paced beats correspond to successful pacemaker-induced depolarization and often exhibit altered QRS morphology. Representative examples are observed in records \#102, \#104, \#107, and \#217. However, these waveform differences reflect pacing-related morphology changes and are not necessarily indicative of pathological arrhythmia.

Atrial Premature Beats (\texttt{A}):
These beats typically exhibit near-normal QRS morphology and are primarily characterized by irregular RR intervals. Consequently, they contribute to false negatives, as the model relies on morphology rather than timing. This behavior is observed in records such as \#100, \#232, and \#209, with more than 200+ FN instances across multiple records. Incorporating lightweight RR interval analysis can improve detection of such beats and support rhythm-level interpretation.


Overall, these observations suggest that combining the proposed ML model with lightweight rule-based features, such as RR interval variability and QRS width estimation, or incorporating additional leads when available, can improve beat-level annotation performance. The proposed approach remains promising for lightweight arrhythmia detection in embedded wearable systems.

\section{Conclusion}
\label{sec_conclusion}
In this work, an optimized and quantized autoencoder framework was developed for ECG arrhythmia detection on ultra-low-power embedded platforms. The proposed lightweight model achieved high recall while maintaining a compact architecture suitable for deployment on resource-constrained microcontrollers. The proposed approach was comprehensively evaluated on the MIT-BIH Arrhythmia Database. The framework is generalizable to other physiological signals, such as EEG or PPG, for local anomaly detection and can be adapted for energy-aware ASIC or FPGA implementations in future biomedical systems.

Deployment on the \textit{ESP32-S3} platform using TensorFlow Lite for Microcontrollers (TFLite Micro) demonstrated the feasibility of end-to-end on-device inference, including data handling, model execution, and anomaly detection without reliance on external computation or hardware accelerators. The system operates within constrained memory and latency budgets, confirming its suitability for continuous ECG monitoring in wearable and edge-based healthcare applications.

The autoencoder-based approach provides an inherent advantage in interpretability, as reconstruction error offers a direct measure of deviation from learned normal patterns. This capability supports transparent anomaly detection and can assist in identifying subtle waveform variations that may not be explicitly labeled in clinical annotations.

Our proposed models achieve reliable anomaly detection performance while maintaining sensitivity to subtle waveform deviations, including cases not explicitly labeled as abnormal. This capability suggests potential usefulness for continuous ECG monitoring systems, where early detection of atypical cardiac activity may support timely diagnosis and intervention, particularly in resource-constrained wearable monitoring systems.

Future work will focus on integrating the proposed model into custom hardware that combines an ECG sensing front end with a dedicated edge-AI inference engine. Further extensions toward multiclass arrhythmia classification, along with optimization through hyperparameter tuning and structured pruning, will be explored to achieve even lower memory footprint and latency. In addition, security and privacy aspects of on-device machine learning, including federated learning pipelines, will be investigated to enhance personalization and support real-world healthcare deployments.

\subsection{System-Level Design Trade-offs}

1) R--R Interval Variability: The current autoencoder model operates on fixed-length ECG segments and does not explicitly capture R--R interval variability. Although R--R variability itself is not an arrhythmia, sustained irregular timing patterns are clinically relevant indicators of rhythm disorders. Consequently, arrhythmias driven primarily by temporal irregularities may be only partially detected at the segment level.

This is particularly relevant for beat types such as atrial premature beats (A), where the primary distinguishing characteristic is irregular timing rather than strong morphological deviation. In such cases, segment-level reconstruction alone may not fully capture the underlying rhythm abnormality.

This limitation can be effectively mitigated through complementary software logic for lightweight R--R interval analysis, where beat-to-beat timing variation is evaluated alongside the segment-level reconstruction error. Such a hybrid approach enables improved rhythm-level interpretation while preserving the computational simplicity required for embedded deployment.

\smallskip

2) Alternating Rhythm Patterns: The observed reduction in classification accuracy is primarily attributed to overlap between Normal and premature beats within R--R segments, particularly in bigeminy and trigeminy rhythms.

For example, Record \#200 consists of a sustained ventricular bigeminy rhythm with 2,792 annotated beats, where Normal and premature ventricular contraction (PVC) beats alternate. Due to this alternating structure, individual R--R segments may contain mixed or transitional morphology, leading to ambiguity at the segment level. As a result, out of 1,743 Normal segments, only 835 were correctly identified by the DNN-based model and 499 by the CNN-based model.

However, from a system-level perspective, this behavior aligns with clinical interpretation. In such rhythms, the diagnostic significance lies not in individual beat classification, but in the persistence of the alternating pattern over time ~\cite{goldberger2017clinical}. Even if some segments are locally classified as normal, the overall rhythm over a longer observation window (e.g., 10--30\,s) clearly indicates an abnormal condition.

Therefore, rather than representing a limitation, this behavior highlights the distinction between segment-level anomaly detection and rhythm-level interpretation. Incorporating lightweight temporal aggregation strategies, such as sliding-window analysis or majority voting across consecutive segments enables robust detection of such repeating arrhythmias in practical deployments.

\smallskip

3) Single-Lead Constraints: The proposed model exhibits limited sensitivity to certain morphology-driven variations, particularly those involving QRS width and shape, such as left and right bundle branch blocks (LBBB and RBBB). Accurate identification of these conditions often depends on multi-lead ECG analysis (for example, inclusion of lead V1 and/or V6), whereas the current implementation relies solely on the MLII lead to support wearable deployment constraints.

As a result, morphology-specific annotations, including L (LBBB) and R (RBBB), may present reduced separability when their distinguishing characteristics are not prominently expressed in the MLII lead. This limitation reflects an inherent trade-off between single-lead wearable ECG systems and the richer spatial information available in clinical 12-lead ECG setups, rather than a limitation of the proposed model.

\subsection{Future Research Directions}
In TinyML settings, quantized models (e.g., INT8) often lose fine-grained detail compared to floating-point (FP32) models due to reduced precision and limited model capacity. As a result, subtle patterns such as small morphological changes or transient events may be suppressed. To address this, techniques such as weighted autoencoders and localized error autoencoders can be employed. By focusing reconstruction error on specific regions (e.g., around the QRS complex) or assigning higher importance to selected potions of the segments, the model can better retain clinically relevant information.

A multi-model approach can further improve performance, where lightweight specialized models are assigned to subsets of classes or tasks. For example, separate models can be used for different ECG leads (e.g., MLII vs.\ V1/V6), along with additional models for tasks such as anomaly detection and arrhythmia classification. This reduces complexity to a single unified model and allows better focus, while keeping each model small and efficient for deployment under resource constraints.

In addition, modified segmentation approaches, such as asymmetric split (30:70) around consecutive R–R intervals, may improve beat-level representation in rhythms with repeating patterns of normal and abnormal beats, better aligning with the conventional ECG cycle and enabling direct beat-level validation using annotated datasets. Finally, extending the current anomaly detection framework toward a multiclass autoencoder architecture could enable classification of specific arrhythmia types and provide richer diagnostic information while remaining suitable for resource-constrained embedded monitoring systems.

\section*{Acknowledgment}
The authors would like to thank Dr.~Ram Naresh Soudri, MBBS, MD, DM, Interventional Cardiologist, for his valuable inputs and clinical guidance provided during the course of this work. The authors acknowledge funding support from the E-Health Research Center (EHRC), IIIT Bangalore.

The authors used AI-assisted tools to improve the language, grammar, and writing style during manuscript preparation. The technical content of the manuscript, including the methodology, experiments, analysis, results, and conclusions, was conceived, validated, and verified by the authors.

\smallskip

\sloppy
\bibliographystyle{IEEEtran}
\bibliography{myContents/references}

\begin{IEEEbiography}
[{\includegraphics[width=1in,height=1.25in,clip,keepaspectratio]{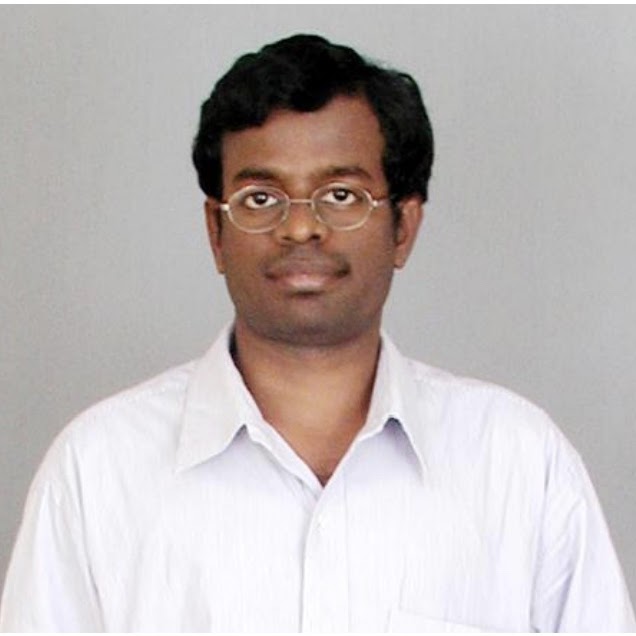}}]{Nagarajan S}
received the M.Tech degree in Embedded Systems and has over 20 years of industry experience in embedded systems and firmware development. He is currently a doctoral researcher at the International Institute of Information Technology Bangalore (IIIT-B), India. His research interests include TinyML, embedded machine learning, real-time signal processing, and edge AI for wearable healthcare systems.
\end{IEEEbiography}


\begin{IEEEbiography}
[{\includegraphics[width=1in,height=1.25in,clip,keepaspectratio]{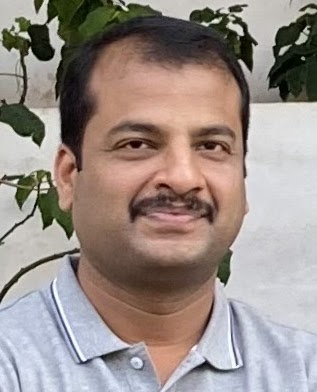}}]{Kurian Polachan}
(Senior Member, IEEE) is an Assistant Professor with the International Institute of Information Technology Bangalore (IIIT-B), India, where he leads the Connected Devices and Wearables Lab (CDWL). He received the M.Tech. and Ph.D. degrees from the Indian Institute of Science (IISc), Bangalore, India. From 2021 to 2022, he was a Postdoctoral Researcher with Purdue University, West Lafayette, IN, USA. Before joining academia, he spent over five years with Cypress Semiconductor, focusing on embedded systems design. His research interests include secure and intelligent wearable systems for next-generation healthcare and IoT applications, with a specific focus on ultra-low-power hardware, embedded intelligence, and hardware-rooted security.
\end{IEEEbiography}

\end{document}